\documentclass[11pt]{article}

\usepackage{url}
\usepackage[latin1]{inputenc}
\usepackage[english]{babel}
\usepackage{graphicx}
\usepackage{amssymb}
\usepackage{amsmath}
\usepackage{amsfonts}
\usepackage{amsthm}
\usepackage{algorithm}
\usepackage{algorithmic}
\usepackage{caption}
\usepackage{subcaption}
\usepackage{color}

\usepackage{setspace}

\newtheorem{remark}{Remark}

\numberwithin{equation}{section}

\,

\title{Sequential Dimensionality Reduction \\ for Extracting Localized Features}
\date{}

\author{
Gabriella Casalino \\ 
Department of Informatics \\ University of Bari ``A. Moro" \\ Via E. Orabona 4, 70125 Bari, Italy \\ 
gabriella.casalino@uniba.it 
\and 
Nicolas Gillis \\ 
Department of Mathematics and Operational Research \\ 
Facult\'e Polytechnique, Universit\'e de Mons \\ 
Rue de Houdain 9, 7000 Mons, Belgium\\
 nicolas.gillis@umons.ac.be  
}

\begin{document}

\maketitle

\begin{abstract} Linear dimensionality reduction techniques are powerful tools for image analysis as they allow the identification of important features in a data set. 
In particular, nonnegative matrix factorization (NMF) has become very popular as it is able to extract sparse, localized and easily interpretable features by imposing an additive combination of nonnegative basis elements. 
Nonnegative matrix underapproximation (NMU) is a closely related technique that has the advantage to identify features sequentially. 
In this paper, we propose a variant of NMU that is particularly well suited for image analysis as it incorporates the spatial information, that is, it takes into account the fact that neighboring pixels are more likely to be contained in the same features, and 
favors the extraction of localized features by looking for sparse basis elements. 
We show that our new approach competes favorably with comparable state-of-the-art techniques on synthetic, facial and hyperspectral image data sets.  
\end{abstract}


\section{Introduction} 
\label{method}

Linear dimensionality reduction (LDR) techniques are powerful tools for the representation and analysis of high dimensional data. 
The most well-known and widely used LDR is principal component analysis (PCA) \cite{J86}.  
When dealing with nonnegative data, it is sometimes crucial to take into account the nonnegativity in the decomposition to be able to interpret the LDR meaningfully. 
For this reason, nonnegative matrix factorization (NMF) was introduced and has been shown to be very useful in several applications such as document classification, air emission control and microarray data analysis; see, e.g.,~\cite{G14} and the references therein. Given a nonnegative input data matrix $M \in \mathbb{R}^{n \times m}_+$ and a factorization rank $r$, 
NMF looks for two matrices $U \in \mathbb{R}^{n \times r}_+$ and $V \in \mathbb{R}^{r \times m}_+$ such that $M \approx UV$. 
Hence each row $M(i,:)$  of the input matrix $M$ is approximated via a linear combination of the rows of $V$: for $1 \leq i \leq n$, 
\[
M(i,:) \quad \approx \quad \sum_{k=1}^r \; U_{ik}  \; V(k,:) .  
\] 
In other words, the rows of $V$ form an approximate basis for the rows of $M$, and the weights needed to reconstruct each row of $M$ are given by the entries of the corresponding row of $U$. 
The advantage of NMF over PCA (that does not impose nonnegativity contraints on the factors $U$ and~$V$) is that the basis elements $V$ can be interpreted in the same way as the data (e.g., as vectors of pixel intensities; see Section~\ref{expres} for some illustrations) 
while the nonnegativity of the weights in $U$ make them easily interpretable as activation coefficients. 
In this paper, we focus on imaging applications and, in particular, on blind hyperspectral unmixing which we describe in the next section.

\subsection{Nonnegative Matrix Factorization for Hyperspectral Images} \label{sec:HyperImages}

\begin{figure}[ht!]
\begin{center}
\[
\hspace{-0.5cm} 
\underbrace{M(i,:)}_{\text{\begin{tabular}{c} spectral signature \\ of $i$th pixel \vspace{0.5cm}\\ 
\includegraphics[width=5cm]{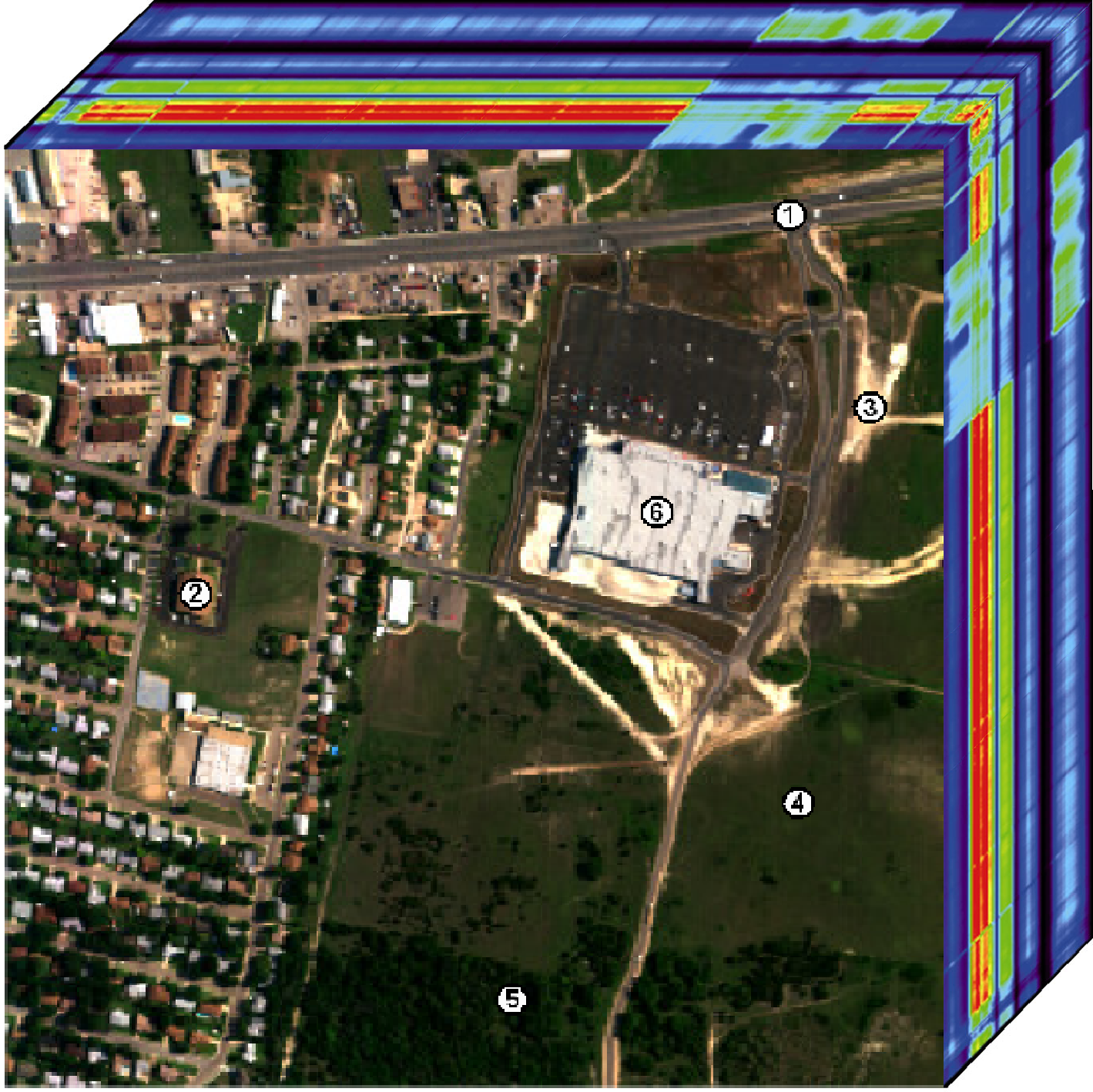} \end{tabular}}} 
\approx   \quad 
\sum_{k=1}^r  
\underbrace{U_{ik}}_{\text{\begin{tabular}{c} abundance of $k$th endmember \\ in $i$th pixel \vspace{0.2cm} \\  
\includegraphics[height=5cm]{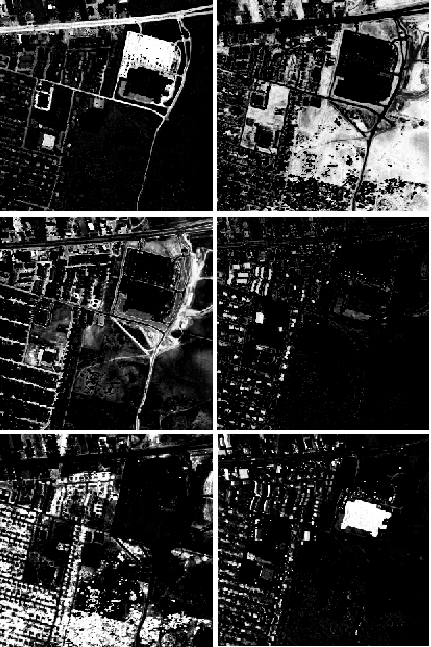} \end{tabular}}} 
\underbrace{V(k,:)}_{\text{\begin{tabular}{c} spectral signature \\ of $k$th endmember \vspace{0.2cm} \\ 
\includegraphics[height=5cm]{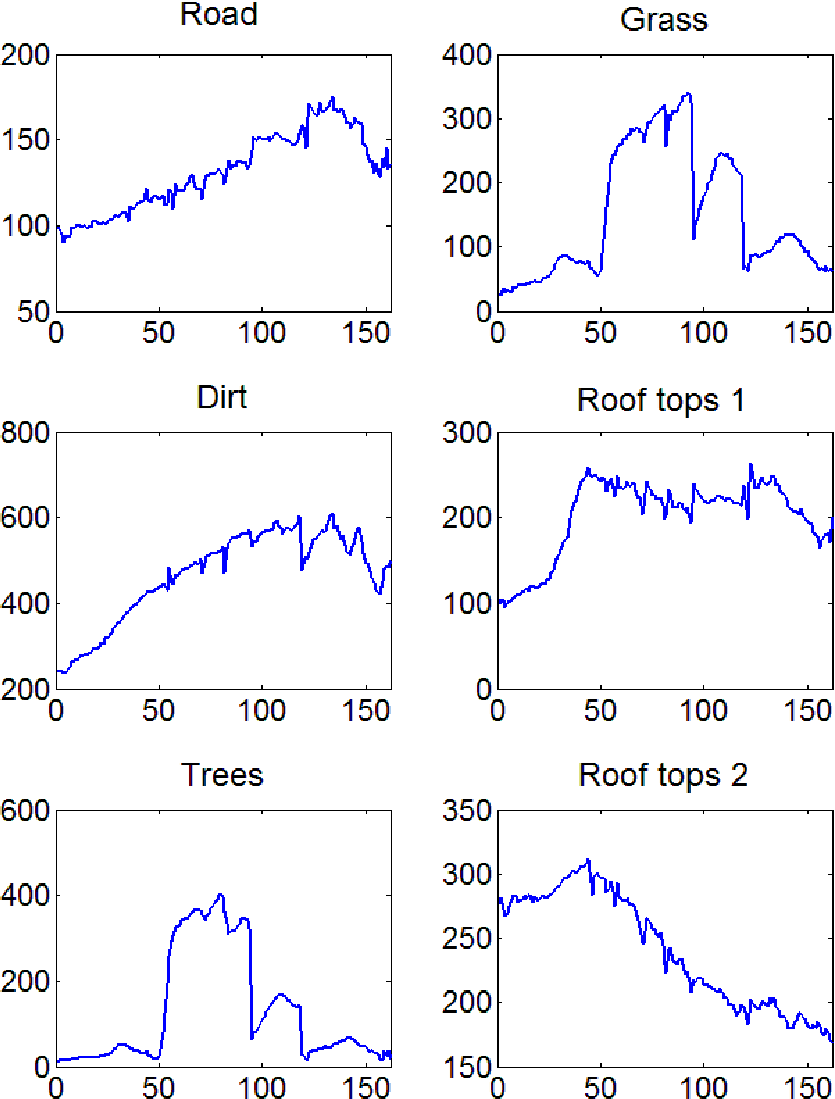} \end{tabular}}}    \hspace{-1cm}.  
\]
\caption{\small Decomposition of the Urban hyperspectral image from \protect\url{http://www.agc.army.mil/}, constituted mainly of six endmembers ($r = 6$): road, grass, dirt, two kind of roof tops and trees. 
Each row of the matrix $V$ is the spectral signature of an endmember, while each row of the matrix $U$ is the abundance map of the corresponding endmember, that is, it contains the abundance of all pixels for that endmember.} 
\label{HyperImage} 
\end{center}
\end{figure}

A hyperspectral image (HSI) is a three dimensional data cube providing the electromagnetic reflectance of a scene at varying wavelengths measured by hyperspectral remote sensors. 
Reflectance varies with wavelength for most materials because energy at certain wavelengths is scattered or absorbed to different degrees, this is referred to as the spectral signature of a material; see, e.g.,~\cite{hyper_image2}.  
Some materials will reflect the light at certain wavelengths, others will absorb it at the same wavelengths. 
This property of hyperspectral images is used to uniquely identify the constitutive materials in a scene, referred to as endmembers, and classify pixels according to the endmembers they contain. 
A hyperspectral data cube can be represented by a two dimensional pixel-by-wavelength matrix 
$M \in \mathbb{R}_{+}^{n\times m}$. 
The columns $M(:,j) \in \mathbb{R}_{+}^{n}$ of $M$ ($1 \leq j \leq m$) are original images that have been converted into $n$-dimensional column vectors (stacking the columns of the image matrix into a single vector), 
while the rows $M(i,:) \in \mathbb{R}_{+}^{m}$ of $M$ ($1 \leq i \leq n$) are the spectral signatures of the pixels (see Figure \ref{HyperImage}). 
Each entry $M_{ij}$ represents the reflectance of the $i$-th pixel at the $j$-th wavelength. 
Under the linear mixing model, the spectral signature of each pixel results from the additive linear combination of the nonnegative spectral signatures of the endmembers it contains. 
In that case, NMF allows to model hyperspectal images because of the nonnegativity of the spectral signatures and the abundances: Given a hyperspectral data cube represented by a two dimensional matrix $M\in\mathbb{R}_{+}^{n\times m}$, NMF approximates it with the product of two factor matrices $U\in\mathbb{R}_{+}^{n\times r}$ and $V\in\mathbb{R}_{+}^{r \times m}$ 
such that the spectral signature of each pixel (a row of matrix $M$) is approximated by the additive linear combination of the spectral signatures of the endmembers (rows of matrix $V$), weighted by coefficients $U_{ik}$ representing the abundance of the $k$-th endmember in the $i$-th pixel. For all $i$, we have: 
\begin{align}
\label{eq: NMF_hyper}
M(i,:) \approx \sum_{k=1}^{r} U_{ik} V(k,:) , 
\end{align}
where $r$ is the number of endmembers in the image. 
The matrix $U$ is called the abundance matrix while the matrix $V$ is the endmember matrix. 
Figure~\ref{HyperImage} illustrates this decomposition on the Urban hyperspectral data cube. 

Unfortunately, as opposed to PCA, NMF is a difficult problem (NP-hard)~\cite{V09}. 
Moreover, the decomposition is in general non-unique and has to be recomputed from scratch when the factorization rank $r$ is modified. For these reasons, a variant of NMF, referred to as nonnegative matrix underapproximation, was recently proposed that allows to compute factors sequentially; it is presented in the next section.

\subsection{Nonnegative Matrix Underapproximation}

Nonnegative matrix underapproximation (NMU)~\cite{GG09} was introduced in order to solve NMF sequentially, that is, to compute one rank-one factor \mbox{$U(:,k)V(k,:)$} at a time: first compute \mbox{$U(:,1)V(1,:)$}, then \mbox{$U(:,2)V(2,:)$}, etc. 
In other words, NMU tries to identify sparse and localized features sequentially.  
In order to keep the nonnegativity in the sequential decomposition, 
it is natural to use the following upper bound constraint for each rank-one factor  of the decomposition: 
for all $1 \leq k \leq r$, 
\[
U(:,k)V(k,:) \leq M - \sum_{p < k} U(:,p)V(p,:)  . 
\]  
Hence, given a data matrix $M\in\mathbb{R}^{n\times m}_+$, 
NMU solves, at the first step, the following optimization problem:
\begin{equation*}
\min_{u\in\mathbb{R}_+^{n}, v\in\mathbb{R}_+^{m}} 
 \left\Vert M-uv^{\mathrm{T}}\right\Vert _{F}^{2} 
\quad \text{such that } \quad  
 uv^{\mathrm{T}}\leq M ,  
\end{equation*}
referred to as rank-one NMU. 
Then, the nonnegative residual matrix $R=M-uv^{\mathrm{T}}\geq0$ is computed, 
and the same procedure can be applied on the residual matrix $R$.  
After $r$ steps, NMU provides a rank-$r$ NMF of the data matrix $M$. 
Compared to NMF, NMU has the following advantages: 
\begin{enumerate}
\item As PCA, the solution is unique (under some mild assumptions) \cite{GP10},
\item As PCA,  the solution can be computed sequentially, and hence the factorization rank does not need to be chosen a priori, and
\item As NMF, it leads to a separation by parts. Moreover the additional underapproximation constraints enhance this property leading to better decompositions into parts~\cite{GG09} (see also Section~\ref{expres} for some numerical experiments). 
\end{enumerate} 
NMU was used successfully for example in hyperspectral~\cite{GP10} and medical imaging~\cite{KCJ11, KC11}, 
and for document classification~\cite{BGG09}.

\subsection{Outline and Contribution of the Paper} \label{sec:outcontr}

Modifications of the original NMU algorithm were made by adding prior information into the model, as it is also often done with NMF; 
see, e.g.,~\cite{CAZP09} and the references therein.  
More precisely, two variants of NMU have been proposed: 
\begin{enumerate}
\item One adding sparsity constraints on the abundance matrix, dubbed sparse NMU~\cite{GP11}, and  

\item One adding spatial information about pixels, dubbed spatial NMU~\cite{PriorsNicolas}. 
\end{enumerate} 
In this paper, we include both sparsity constraints and spatial information in NMU. 
This allows us to extract localized features in images more effectively. 
We present our algorithm in Section~\ref{newalgo} and show in Section~\ref{expres} that it competes favorably with comparable state-of-the-art techniques (NMF, sparse NMF and several variants of NMU) on synthetic, hyperspectral and facial image data sets.

\section{Algorithm for NMU with Priors} \label{newalgo}

In this section, we describe our proposed technique that will incorporate both spatial and sparsity priors into the NMU model. This will allow us to extract more localized and more coherent features in images. 

\subsection{Reformulation of NMU and Lagrangian-Based Algorithm}  

First, let us briefly describe the original NMU algorithm from \cite{GG09}. 
The original rank-one NMU problem is the following 
\begin{equation}
\min_{u\in\mathbb{R}_+^{n}, v\in\mathbb{R}_+^{m}} 
 \left\Vert M-uv^{\mathrm{T}}\right\Vert _{F}^{2} 
\quad \text{such that } \quad  
 uv^{\mathrm{T}}\leq M . 
\label{NMU}
\end{equation}  
As described in the introduction, solving this problem allows to compute NMF sequentially, that is, one rank-one factor at a time while preserving nonnegativity. 
In~\cite{GG09}, approximate solutions for rank-one NMU~\eqref{NMU} are obtained using the Lagrangian dual
\begin{align}
\max_{\Lambda\geq0}\min_{u\geq0,v\geq0} L\left(u,v,\varLambda\right)
= \max_{\Lambda\geq0}\min_{u\geq0,v\geq0} 
\left\Vert M-uv^{\mathrm{T}}\right\Vert _{F}^{2}+2\sum_{i,j}\left(uv^{\mathrm{T}}-M\right)_{ij}\varLambda_{ij} , 
\label{eq:lagrangian}
\end{align}
where $\varLambda\in\mathbb{R}^{n \times m}$ is the matrix containing the Lagrangian multipliers of the underapproximation constraints.
The authors prove that for a fixed $\varLambda$, the problem $\min_{u\geq0,v\geq0}L\left(u,v,\varLambda\right)
$, called Lagrangian Relaxation of \eqref{NMU}, is equivalent to 
$\min_{u\geq0,v\geq0} \left\Vert (M-\Lambda) - uv^{\mathrm{T}}\right\Vert _{F}^{2}$ which is equivalent, up to a scaling of the variables, to 
\begin{align} 
\max_{u\geq0, v\geq0}u^{\mathrm{T}}\left(M-\varLambda\right)v 
\quad \text{ such that} \quad  
\left\Vert u\right\Vert_{2} \leq 1, \left\Vert v\right\Vert_{2} \leq 1  . 
\label{eq:lageq}
\end{align}
In fact, one can show that any optimal solution of \eqref{eq:lagrangian} is, 
up to a scaling factor, an optimal solution of \eqref{eq:lageq}, and vice versa. 
It has to be noted that, except for the trivial stationary point (that is, $u = 0$ and $v= 0$, which is optimal only for $M - \Lambda \leq 0$), 
any stationary point of \eqref{eq:lageq} will satisfy $||u||_2 = ||v||_2 = 1$ (this can be shown by contradiction). Therefore, without loss of generality, one can assume $\left\Vert u\right\Vert_{2} = 1$ and $\left\Vert v\right\Vert_{2} = 1$. 

Based on these observations, the original NMU algorithm optimizes alternatively over the variables $u$ and $v$, and updates the Lagrangian multipliers accordingly. 
The advantage of this scheme is that it is relatively simple as the optimal solution of $u$ given $v$ can be written in closed form (and vice versa). The original NMU algorithm iterates between the following steps: 
\begin{itemize}
\item 
$u \leftarrow \max \left( 0 , {(M - \Lambda) v} \right)$, $u \leftarrow u / ||u||_2$ , 

\item
$v \leftarrow \max \left( 0 , {(M - \Lambda)^T u} \right)$, $v \leftarrow v / ||v||_2$  , 

\item
$\Lambda \leftarrow \max \left( 0 , \Lambda + \mu (\sigma uv^T - M) \right)$ where $\sigma = u^T (M - \Lambda) v$ and $\mu$ is a parameter. 
\end{itemize}
Note that given $u$ and $v$, $\sigma = u^T (M - \Lambda) v$ is the optimal rescaling of the rank-one factor $uv^T$ 
to approximate $M - \Lambda$. The parameter $\mu$ can for example be equal to $\frac{1}{t}$ where $t$ is the iteration index to guarantee convergence~\cite{GG09}.  
Note also that the original NMU algorithm shares similarities with the power method used to compute the best rank-one unconstrained approximation; see the discussion in~\cite{GP10}.

\subsection{Spatial Information}
\label{Spatial}

The first information that we incorporate in NMU is that 
neighboring pixels are more likely to be contained in the same features.  
The addition of spatial information into NMU improves the decomposition of images by generating spatially coherent features (e.g., it is in general not desirable that features contain isolated pixels). 
In this paper, we use the anisotropic total variation (TV) regularization; see, e.g.,~\cite{IBP11}. 
We define the neighborhood of a pixel as its four adjacent pixels (above, below, left and right). 
To evaluate the spatial coherence, we use the following function 
\begin{align}
\label{reg_local1}
\sum_{i=1}^{n}\sum_{j\in \mathcal{N}\left(i\right)}\left|u_{i}-u_{j}\right|=2\left\Vert Nu\right\Vert _{1} , 
\end{align}
where $\mathcal{N}\left(i\right)$ is the set of neighboring pixels of pixel $i$, and $N\in\mathbb{R}^{K\times n}$ is a neighbor matrix
where each column corresponds to a pixel and each row indicates a pair $\left( i,j \right)$ of neighboring pixels: 
\begin{align}
\label{neighborgMatrix}
N\left(k,i\right)=1\quad \text{and} \quad N(k,j)=-1 , 
\end{align}
with $1\leq i<  j \leq m$ and $j \in \mathcal{N}\left(i\right)$, and $K$ is the number of neighboring pairs ($K \leq 4n$, because each pixel has at most 4 neighbors). The term $||Nu||_1$ therefore accounts for the distances between each pixel and its neighbors. 
Note that the $\ell_1$ norm is used here because it is able to preserve the edges in the images, as opposed to the $\ell_2$ norm which would smooth them out; see, e.g.,~\cite{ZKZ07, IBP11}. 
The same penalty was used in~\cite{PriorsNicolas} to incorporate spatial information into NMU.

\subsection{Sparsity Information}
\label{Sparsity}

The second information incorporated in the proposed algorithm is that each feature should contain a relatively small number of pixels. In hyperspectral imaging, this translates into the fact that each pixel usually contains only a few endmembers: the row vectors of the abundance matrix $U$ should have only few non-zero elements. 
In~\cite{GP11}, the authors demonstrated that incorporating this sparsity prior into NMU leads to better decompositions. 
They added a regularization term to the objective function in \eqref{NMU} based on the $\ell_1$-norm heuristic approach, in order to minimize the non-zero entries of $u$: $\left\Vert u\right\Vert _{1}$ where $\left\Vert u\right\Vert _{2}=1$. 
This is the most standard approach to incorporate sparsity and was also used to design sparse NMF algorithms~\cite{Kim}.

\subsection{Algorithm for NMU with Spatial and Sparsity Constraints} \label{sec:Lagrangian}

In order to inject information in the factorization process, we add the regularization terms for the sparsity and spatial information in the  NMU formulation~\eqref{eq:lageq}: 
\begin{equation}
\max_{\left\Vert u\right\Vert _{2}\leq1,\left\Vert v\right\Vert _{2}=1, u\geq0,v\geq0} 
\left(\underbrace{    u^{\mathrm{T}}\left(M-\Lambda\right)v-\varphi\left\Vert u\right\Vert _{1}-\mu\left\Vert Nu\right\Vert _{1}}_{\textmd{  $=f\left(u,v \right)$ }}\right) . 
\label{LagrangianRelaxation}
\end{equation} 
The objective function is composed of three terms: the first one relates to the classical least squares residual, 
the second enhances sparsity of the abundance vector $u$ while the third improves its spatial coherence. 
The regularization parameters $\mu$ and $\varphi$ are used to balance the influence of the three terms. 
We will refer to \eqref{LagrangianRelaxation} as Prior NMU (PNMU). 
Note that we relax the constraint $||u||_2 = 1$ from NMU to $||u||_2 \leq 1$ to have a convex optimization problem in $u$; see below for more details.

Algorithm~\ref{SNMUalgo} formally describes the alternating scheme to solve the NMU problem with sparsity and spatial constraints \eqref{LagrangianRelaxation}. 
\algsetup{indent=1em}
\begin{algorithm}[ht!]
\caption{Prior NMU: incorporating spatial information and sparsity into NMU}  \label{SNMUalgo}
\begin{algorithmic}[1]
\REQUIRE $M \in \mathbb{R}^{n \times m}_+$, $r \in \mathbb{N}_{+}$, 
$0 \leq \varphi' \leq 1$,  
$0 \leq \mu' \leq 1$,  
small parameter $\epsilon>0$, 
maxiter, iter.
\ENSURE $(U,V) \in \mathbb{R}^{n \times r}_+ \times \mathbb{R}^{r \times m}_+$ s.t. $UV \leq M$ with $U$ containing sparseness and locality information.\\

\STATE Generate the matrix $N$ according to \eqref{neighborgMatrix};
\FOR {$k \, = \, 1$ : $r$}  
	\STATE $[x,y,\varLambda] =$  rank-one underapproximation($M$);   \% Initialization of $\left(x,y\right)$ with an approximate solution to NMU (\ref{NMU}) \label{xyl-initialization}
	\STATE $u_{k}\leftarrow x; \, v_{k}\leftarrow y; \, x\leftarrow\frac{x}{\left\Vert x\right\Vert _{2}}; \, y\leftarrow\frac{y}{\left\Vert y\right\Vert _{2}};$
         \STATE $w_{i}=\left(\left|Nx\right|_{i}+\epsilon\right)^{-0.5}; \; W=\text{diag}(w);$  \% Initialization of IRWLS weights
	\STATE $\varphi =\varphi' \left\Vert \left(M-\varLambda\right)y\right\Vert _{\infty}$ \% Setting the sparsity parameter $\varphi$ \label{phi}
	\STATE $x=\max\left(0,\left(x- \varphi \right)\right); \, x=\frac{x}{\left\Vert x\right\Vert _{2}};$ 
	\STATE $z=\text{rand}(m,1)$;  \% Estimate of the eigenvector of $B$ associated with the largest eigenvalue 
	\FOR {$t \, = \, 1$ : maxiter} 
		\STATE $A=M-\Lambda;$
		\STATE \% Update of $x$ 
		\STATE $B=\left(WN\right)^{\mathrm{T}}\left(WN\right);$ \label{B} \label{xupdate_s}
    		\STATE for $l=1$ : iter $z=Bz;$ $z=\frac{z}{\left\Vert z\right\Vert _{2}};$  \%Power method  \label{Power}
		\FOR{$l \, = \, 1$ : iter} 
			\STATE $\mu=\mu' \frac{\left\Vert Ay\right\Vert _{\infty}}{\left\Vert Bx\right\Vert _{\infty}};$ \label{mu} \% Setting of the spatial parameter $\mu$
                           \STATE $L=\max\left(\varepsilon,\mu\left(z^{\mathrm{T}}Bz\right)\right)$ \% Approximated Lipschitz constant
	         	\STATE $\nabla f(x)=Ay-\mu Bx-\varphi;$ \label{gradient} 
			\STATE $x\leftarrow\mathcal{P}\left(Lx+\nabla f\left(x\right)\right);$
		\ENDFOR \label{xupdate_e}
		\STATE \% Update of $y$ 
		\STATE $y\leftarrow\max\left(0,A^{\mathrm{T}}x\right);$ \label{yupdate_s}
                  \STATE if $\left\Vert y\right\Vert _{2}\neq0$ then $y\leftarrow\frac{y}{\left\Vert y\right\Vert _{2}};$ \label{yupdate_e}
		\STATE \%Update of $\Lambda$ and save $\left(x,y\right)$ 
		\IF {$x\neq0$ \, and \, $y\neq0$} \label{lambdaupdate_s}
			\STATE $\sigma=x^{\mathrm{T}}Ay;$ $u_{k}\leftarrow x; \;  v_{k}\leftarrow\sigma y;$
			\STATE $\Lambda\leftarrow\max\left(0,\Lambda-\frac{1}{t+1}\left(M-u_{k}v_{k}^{\mathrm{T}}\right)\right);$
		\ELSE
			\STATE $\Lambda\leftarrow\frac{\Lambda}{2};$
			\STATE $x\leftarrow\frac{u_{k}}{\left\Vert u_{k}\right\Vert _{2}}; \;  y\leftarrow\frac{v_{k}}{\left\Vert v_{k}\right\Vert _{2}};$
		\ENDIF \label{lambdaupdate_e}
		\STATE \% Update of the weights
		\STATE $w_{i}=\left(\left|Nx\right|_{i}+\epsilon\right)^{-0.5}; \; W=\text{diag}(w);$ \label{W_i}
	\ENDFOR
	\STATE $M=\max\left(0,M-u_{k}v_{k}^{\mathrm{T}}\right);$
\ENDFOR
\end{algorithmic}
\end{algorithm}

As in the original NMU model, the optimal solution for $v$ given $u$ can still be written in closed form. However, because of the spatial information (the term $||Nu||_1$), there is no closed form for the optimal solution of $u$ given $v$, although the problem is convex in $u$. 
In order to find an approximate solution to that subproblem, 
we combine iterative reweighted least squares with a standard projected gradient scheme from~\cite{Y04}, 
as proposed in~\cite{PriorsNicolas}; see below for more details.  
Finally, a simple block-coordinate descent scheme is used to find good solutions to problem (\ref{LagrangianRelaxation}). This is achieved by applying the following alternating scheme that optimizes one block of variables while keeping the other fixed:
\begin{enumerate}
\item 
$v \leftarrow \max\left(\left(M-\varLambda\right)^{\mathrm{T}}u,0\right)$ , $v \leftarrow v / ||v||_2$ (lines \ref{yupdate_s} and \ref{yupdate_e})

\item $u\leftarrow \mathcal{P}\left(u - \frac{1}{L} \nabla f_w\left(u\right)\right) $ 
(from line \ref{xupdate_s} to line \ref{xupdate_e}) 
\\ 
where $L$ is the Lipschitz constant of $\nabla f_w\left(u\right)$, where $f_w$ is a smooth approximation of $f$; see the paragraph below and Equation~\eqref{fw}. 

The projection is defined as 
\\  
$\mathcal{P}\left(s\right)=\begin{cases}
\frac{\max\left(0,s\right)}{\left\Vert \max\left(0,s\right)\right\Vert _{2}} & \text{if}\left\Vert \max\left(0,s\right)\right\Vert _{2}\geq1, \\
\max\left(0,s\right) & \text{otherwise.} 
\end{cases}
$
\item $\varLambda\leftarrow\max\left(0,\varLambda + \mu \left(\sigma uv^{\mathrm{T}}-M\right)\right)$ with $\sigma = u^{\mathrm{T}} M v$  
(from line \ref{lambdaupdate_s} to line \ref{lambdaupdate_e}).  
\end{enumerate}
Variables $\left(u,v,\varLambda\right)$ are initialized with a solution of the original NMU algorithm from~\cite{GP11} 
(line \ref{xyl-initialization}).

\paragraph{Smooth approximation of $f$} The variables $v$ and $\varLambda$ are updated as in the original NMU algorithm, whilst the update of $u$ is modified in order to take into account the penalty terms. 
The update of $u$ involves the non-differentiable $\ell_1$-norm term $||Nu||_1$ in the objective function $f$. 
Note that, since $u \geq 0$, $||u||_1 = \sum_{i=1}^m u_i$ hence it is differentiable on the feasible set. 
The authors in~\cite{PriorsNicolas} suggested to use iteratively re-weighted least squares (IRWLS) to approximate the $\ell_1$-norm. 
At the $t$-th iteration, the term $\left\Vert Nu\right\Vert _{1}$ is replaced with 
\begin{equation}
\left\Vert Nu\right\Vert _{1}\approx u^{\mathrm{T}}\left( \underbrace{ N^{\mathrm{T}}W^{\left(t\right)\mathrm{T}}W^{\left(t\right)}N}_{\textmd{$=B$}} \right)u , 
\end{equation} 
where $W^{\left(t\right)}=\text{diag}\left(w^{\left(t\right)}\right)$, 
diag($x$) takes a vector $x$ as an input and returns a diagonal matrix with the entries of $x$ on the diagonal,
 and 
$w_{i}^{\left(t\right)}
=\left(\left|Nu^{\left(t - 1\right)}\right|_{i}+\varepsilon\right)^{-\frac{1}{2}}$ 
(lines \ref{B} and \ref{W_i}). 
The idea is to approximate the $\ell_1$-norm of a vector $z \in \mathbb{R}^n$ as a weighted $\ell_2$-norm: 
\[
|| z ||_1 \approx \sum_{i=1}^n w_i z_i^2 , \quad \text{with } w_i = \frac{1}{ |z_i| + \epsilon} , 
\]
where $\epsilon$ is a small constant (we used $10^{-3}$). 
In our case, the weights are estimated using the value of $u$ at the previous iteration. We denote $f_w$ the modified objective function that approximates $f$ at the current iterate.   
Hence the non-differentiable term $||Nu||_1$ is approximated with a convex and quadratic differentiable term $u^T B u$, 
and we can use a standard gradient descent scheme from smooth convex optimization~\cite{Y04} (line \ref{gradient}); the gradient being 
\begin{equation} \label{fw} 
\nabla f_w \left(u\right)=\left(M-\varLambda\right)v-\varphi e-\mu\left(Bu\right) , 
\end{equation} 
where $e$ is the vector of all ones of appropriate dimension. 
The Lipschitz constant $L$ of $\nabla f_w$ is equal to the largest singular value of $B$. 
 Since the matrix $B$ is rather large (but sparse), it is computational costly to  compute exactly its largest eigenvalue, hence we use several steps of the power method (line \ref{Power}) to estimate the Lipschitz constant $L$.

\paragraph{Choice of the penalty parameters $\varphi$ and $\mu$} 
It is in general difficult to choose a priori `optimal' values for the penalty parameters 
$\varphi$ (sparsity parameter) and $\mu$ (local information). 
In fact, it is difficult to know the sparsity and spatial coherence of the unknown localized features. Moreover, it is difficult to relate them to the parameters $\varphi$ and $\mu$. Note that this choice also depends on the scaling of the input matrix: if $M$ is multiplied by a constant, the first term in \eqref{LagrangianRelaxation} is increased by the same constant.  

In this paper, we use scaling-independent parameters: we replace $\varphi$ and $\mu$ with (lines~\ref{mu} and~\ref{phi}):
\[ 
\mu=\mu' \frac{\left\Vert (M-\varLambda) v\right\Vert _{\infty}}{\left\Vert Bu\right\Vert _{\infty}}
\quad \text{ and } \quad
\varphi=\varphi' \left\Vert \left(M-\varLambda\right)v\right\Vert _{\infty} , 
\] 
for some $\mu' \in [0,1]$ and  $\varphi' \in [0,1]$. 
These choices were proposed respectively in~\cite{PriorsNicolas} and~\cite{GP11}. 
They have the advantage to give a range of possible values for the penalty parameters: 
\begin{itemize}

\item $\varphi' = 1$ is the highest possible sparsity level: it would imply that the largest value of $x$ is set to zero by the thresholding operator in the projection $\mathcal{P}$.  

\item $\mu' = 1$ would give more importance to the spatial coherence and will generate extremely smooth features (see next section for some numerical experiments). 

\end{itemize} 

\paragraph{Computational Cost} The computational cost of Algorithm~\ref{SNMUalgo} is the same as the other NMU variants: 
it requires $O(mn)$ operations to compute each rank-one factor, for a total of $O(mnr)$ operations. 
Most of the cost resides in the matrix-vector products ($Av$ and $A^Tu$ for the updates of $u$ and $v$) and the update of the Lagrangian multipliers $\Lambda$. 
Hence, the cost is linear in $m$, $n$ and $r$, as for most NMF algorithms (in particular the one described in the next section).

\paragraph{Convergence} The convergence analysis of Algorithm~\ref{SNMUalgo} is nontrivial, because it combines several strategies that are themselves difficult to analyze. In fact, our algorithm is based on a Langragian relaxation (whose subproblems are solved using a two-block coordinate descent method) where the Lagrangian variables $\Lambda$ are updated to guarantee the convergence of the scheme~\cite{AW07} (see also the discussion in~\cite{GG09}). 
This is combined with a reweighted least squares approach to approximate the non-differentiable $\ell_1$ norm $||Nu||_1$, which is also guaranteed to converge~\cite{DDF10}. In practice, we have observed that Algorithm~\ref{SNMUalgo} converges usually within 500 iterations (but this depends in particular on the size of the data set); see Section~\ref{synt} for some numerical experiments.

\section{Experimental Results} \label{expres}

In this section, we conduct several experiments to show the effectiveness of PNMU. 

In the first part, we use synthetic data sets for which we know the ground truth hence allowing us to quantify very precisely the quality of the solutions obtained by the different algorithms.  

In the second part, we validate the good performance of PNMU on real data sets, and compare the algorithms on two widely used data sets, namely, the Cuprite hyperspectral image and the CBCL face data set. 

The following algorithms will be compared:
\begin{itemize}

\item {\bf Nonnegative matrix factorization (NMF)}. We use the accelerated HALS algorithm (A-HALS)~\cite{GG11} (with parameters $\alpha=0.5$ and $\varepsilon=0.1$, as suggested by the authors). 

\item {\bf Sparse NMF (SNMF)}. It is a sparse version of A-HALS~\cite{G12} that adds an $\ell_1$-norm sparsity-enhancing term for the abundance matrix $U$. We run SNMF with target sparsity for the matrix $U$ given by the sparsity of the abundance matrix $U$ obtained by PNMU. 
Hence we can compare PNMU with SNMF meaningfully as they will have comparable sparsity levels, as it was done in~\cite{GG09} to compare NMU with SNMF. 

\item {\bf Nonnegative matrix underapproximation (NMU)}. This is PNMU with $\varphi'=0$ and $\mu'=0$. 

\item {\bf Nonnegative matrix underapproximation with local information (LNMU)}.  
This is PNMU with $\varphi'=0$. 

\item {\bf Sparse nonnegative matrix underapproximation (SNMU)}.  
This is PNMU with $\mu'=0$.

\item {\bf Nonnegative matrix underapproximation with prior information (PNMU)}. This is Algorithm~\ref{SNMUalgo}. 

\end{itemize}

All the numerical results are obtained by implementing the algorithms in Matlab 7.8 codes and running them on a machine equipped with an Intel\textsuperscript{\textregistered} Xeron\textsuperscript{\textregistered} CPU E5420 Dual Core 250 GHz, RAM 8.00 GB.
The code is available online from~\url{https://sites.google.com/site/nicolasgillis/code}. We use maxiter = 500, iter = 10 for all experiments in the paper.

\subsection{Synthetic data sets} \label{synt}

\begin{figure*}[ht!]
\begin{center}
\includegraphics[width=9cm]{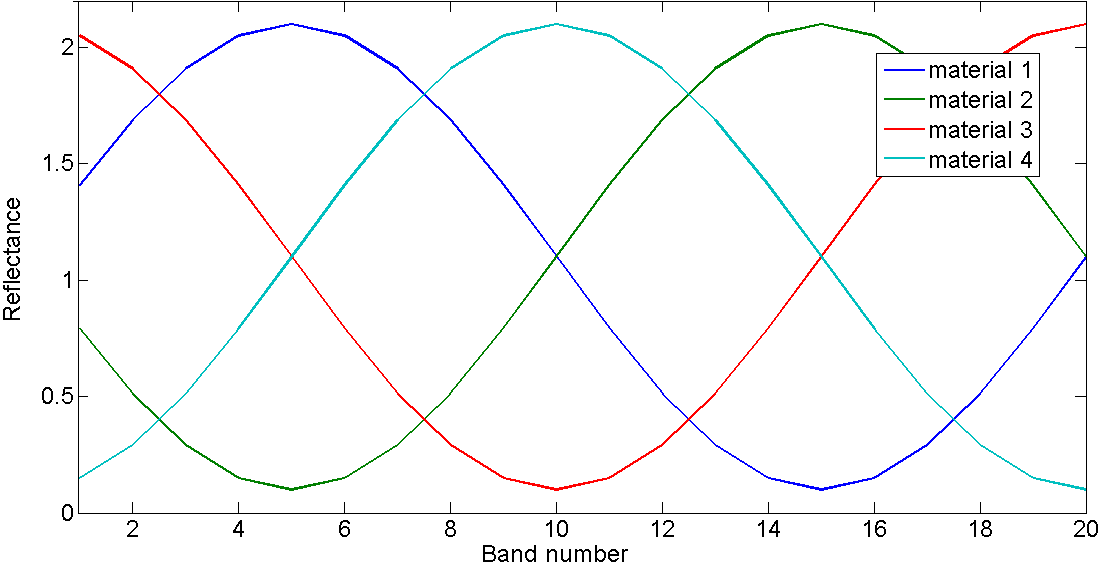} 
\caption{Synhetic spectral signatures.}  
\label{ssignature}
\end{center}
\end{figure*}

Let us construct synthetic hyperspectral images. We use $r=4$ materials perfectly separated. Materials are adjacent rectangles of size $10 \times (k+1)$ pixels with $k=1,2,3,4$, so that the image at each wavelength contains $10 \times 14$ pixels ($m = 140$); 
see the top row of Figure~\ref{syntexres}. 
Mathematically, for $1 \leq i \leq 140$ and $1 \leq k \leq r$, 
\[
U(i,k) = \left\{ \begin{array}{cc}
1 & \text{if }  5(k-1)(k+2) + 1 \leq i \leq 5(k-1)(k+2) + 10 (k+1) , \\
0 & \text{otherwise.} \\
\end{array} \right.  
\] 
We use $n=20$ wavelengths. The spectral signatures are sinusoids with different phases plus a constant to make them nonnegative: 
\[
V(j,k) = 1.1 + \sin \left( x(j) + \phi(k) \right) \; \geq \; 0, 
\]
where $x(j) = j \, \frac{2 \pi}{n}$ with $j = 1,2,\dots,n$, 
and $\phi(k) = (k-1) \frac{2 \pi}{r}$ with $k = 1,2,\dots,r$. 
We also permute the rows of $V$ so that adjacent materials have less similar spectral signatures 
(we used the permutation $[1 \, 3 \, 2 \, 4]$).  
Figure~\ref{ssignature} shows the spectral signatures. 

For the noise, we combine Gaussian and salt-and-pepper  noise. Let us denote $\bar{M}=1.1$ the average of the entries of $UV$ (which is the noiseless synthetic HSI): 
\begin{itemize}
\item Gaussian. We use 
\[
G(i,j) \sim g \, \bar{M} \, \mathcal{N}(0,1), \; 1 \leq i \leq n, 1 \leq j \leq m, 
\]
where $g$ is the intensity of the noise. We use the  function \texttt{randn(n,m)} of Matlab. 

\item Salt-Pepper. For the salt-and-pepper noise, we use a sparse matrix $P$ of density $p$ whose non-zero entries are distributed following $\bar{M} \, \mathcal{N}(0,1)$. We use the function \texttt{sprandn(n,m,p)} of Matlab. 

\end{itemize}

\begin{figure*}[ht!]
\begin{center}
\includegraphics[width=8cm]{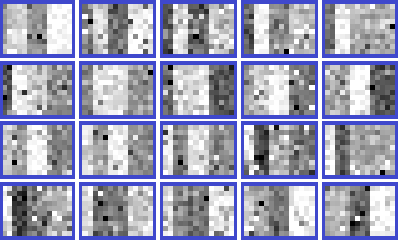} 
\caption{Example of a synthetic HSI with $g = 30\%$ and $p=15\%$. Each image corresponds to a wavelength.}  
\label{syntexample}
\end{center}
\end{figure*}

Finally, we generate the noisy hyperspectral image $M = UV + G + P$; see Figure~\ref{syntexample} for an example. 
Given a solution $(\tilde{U}, \tilde{V})$ generated by an algorithm using matrix $M$, 
we will compare the quality of the $r$ extracted materials as follows. First, we normalize the columns of $\tilde{U}$ so that the maximal entries are equal to one, that is, 
$\tilde{U}(:,k) \leftarrow \frac{\tilde{U}(:,k)}{\max_i \tilde{U}(i,k)}$ for all $k$. Then, we permute the columns of $\tilde{U}$ in order to minimize 
\begin{equation} \label{match}
\text{match}(U, \tilde{U}) = \frac{1}{m r} 
\sum_{k=1}^r ||U(:,k) - \tilde{U}(:,k)||_2^2 \; \in \; [0,1]. 
\end{equation}
The match will be equal to zero if $\tilde{U}$ coincides with $U$ (up to a permutation of the columns), and equal to 1 when $U(i,k) = 0 \iff  \tilde{U}(i,k) = 1$ for all $i,k$ (perfect mismatch). 

Note that for our particular $U$, the match value of the zero matrix is equal to $\text{match}(U, 0)=25\%$ (which is the density of $U$) hence we should not expect values higher than that in our numerical experiments. 
As we will see, match values higher than 2\% correspond to a relatively poor recovery of the matrix $U$; see Figure~\ref{syntexres}.

\subsubsection{Choice of $\phi'$ and $\mu'$}

\begin{figure*}[ht!]
\begin{center}
\includegraphics[width=12cm]{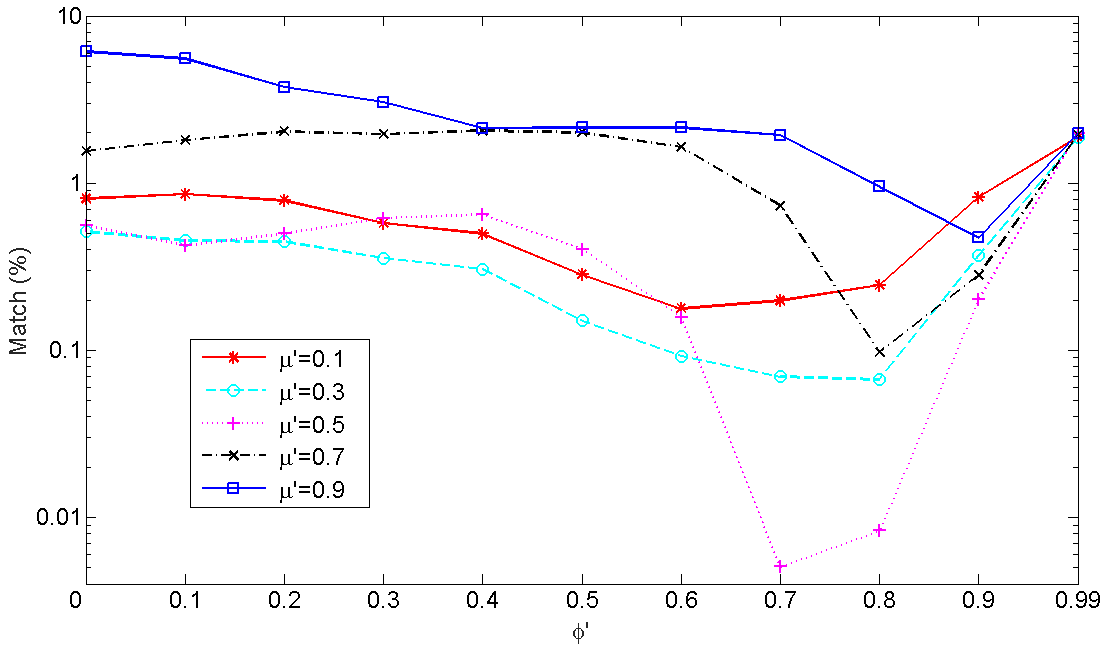} 
\caption{Average value of the match~\eqref{match} of PNMU over 20 randomly generated synthetic HSI with $g=0.2$ and $p=0.05$ 
for PNMU for different values of $\phi'$ and $\mu'$.}  
\label{synt1exp}
\end{center}
\end{figure*}

First, let us observe the influence of $\phi'$ and $\mu'$ on PNMU and choose reasonable values for these parameters for these synthetic data sets. We fix the noise level to $g = 0.2$ and $p=0.05$, and vary $\phi'$ and $\mu'$. 
Figure~\ref{synt1exp} shows the average match of PNMU over 20 randomly generated synthetic data sets for different values of $\phi'$ and $\mu'$. 

\begin{figure*}[ht!]
\begin{center}
\includegraphics[width=6cm]{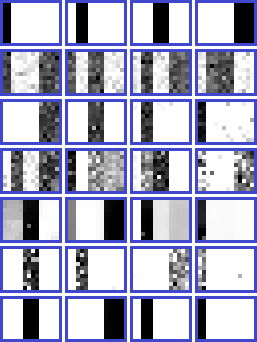} 
\caption{Basis elements extracted by the different algorithms for the synthetic HSI with $g = 0.3$ and $p=0.15$ of Figure~\ref{syntexample}. From top to bottom: True materials, NMF, SNMF with sparsity 0.75, NMU, LNMU, SNMU and PNMU with $\phi' = 0.8$ and $\mu' = 0.5$. The match values are: 
(NMF) 14.34\%, 
(SNMF) 1.64\%, 
(NMU)  13.17\%, 
(LNMU)  2.12\%, 
(SNMU) 7.26\%, 
(PNMU) 0.003\%. 
 }  
\label{syntexres}
\end{center}
\end{figure*}

We observe that values of $\phi' \in [0.6,0.9]$ and $\mu' \in [0.1,0.5]$ provide good values for the match (below 1\%, which is better than the other NMU and NMF algorithms; see Figure~\ref{syntexres}).   
For the comprehensive comparison of the different algorithms in the next section, 
we will use $\phi' = 0.7$ and $\mu' = 0.5$ for LNMU, SNMU and PNMU. 

Figure~\ref{syntexres} shows the materials extracted by 
NMU, NMF, SNMF (with sparsity 0.75), LNMU, SNMU and PNMU with $\phi' = 0.7$ and $\mu' = 0.5$ for the noise level $g = 0.3$ and $p=0.15$. 
We observe that PNMU identifies perfectly the four materials.  
SNMU performs relatively well but returns the basis elements with salt-and-pepper noise. 
LNMU returns spatially more coherent basis elements but they are mixture of several materials (hence not sparse enough).  
NMF and SNMF both return spatially less coherent basis elements, and SNMF returns more localized (sparser) ones identifying relatively well the materials (except for the smallest one).  
\begin{remark} 
The code to generate these synthetic HSI, 
and compare the different algorithms is available from \url{https://sites.google.com/site/nicolasgillis/}. Moreover, one can easily change the noise level, the number and the size of the materials, and the number of wavelengths. The values $g = 0.3$ and $p=0.15$ were chosen because they correspond to a high noise level where PNMU still performs perfectly (see Figure~\ref{gnpsnoise}). As we will see below, PNMU outperforms the other approaches for most reasonable values of the noise level. 
\end{remark}

\begin{figure*}[ht!]
\begin{center}
\includegraphics[width=11cm]{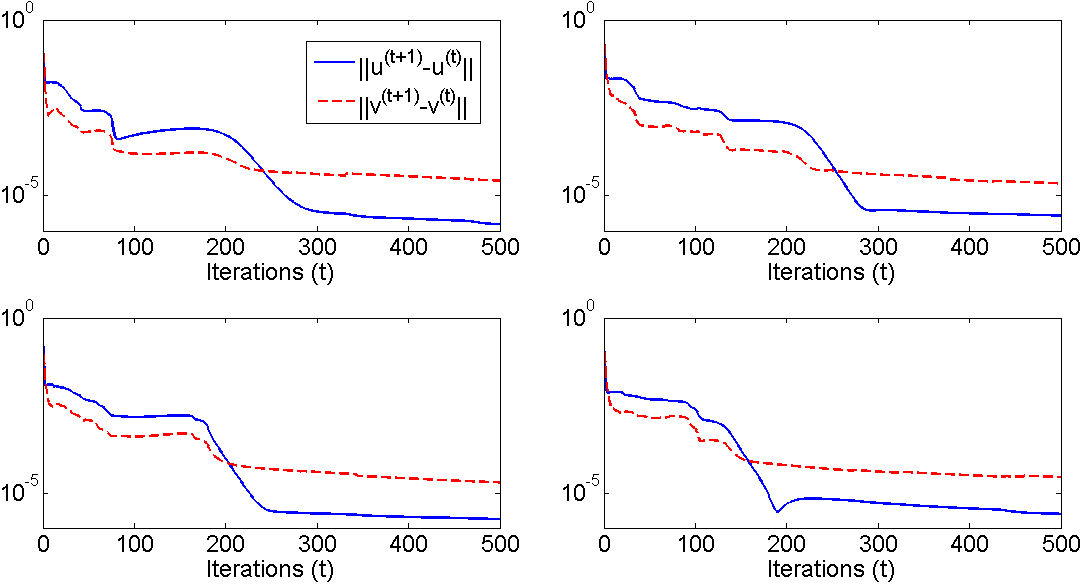} 
\caption{Convergence for each rank-one factor generated by PNMU on the synthetic HSI with $g = 0.3$ and $p=0.15$ 
from Figure~\ref{syntexample}. 
Each plot displays the evolution of $||u^{(t+1)}-u^{(t)}||_2$ and $||v^{(t+1)}-v^{(t)}||_2$ where $(u^{(t)},v^{(t)})$ is the $t$th iterate generated by PNMU for each rank-one factor $uv^T$ computed sequentially  (from left to right, top to bottom).} 
\label{convsynth} 
\end{center}
\end{figure*}

Regarding the convergence of PNMU, although it is difficult to analyze rigorously (cf.\@ the discussion in the previous section), 
Figure~\ref{convsynth} displays the difference between iterates for this particular numerical experiment, showing the relative fast convergence in this case. 

\subsubsection{Comparison}

\begin{figure}
 	\centering
 	\begin{subfigure}[b]{0.9\textwidth}
 		\includegraphics[width=0.9\textwidth]{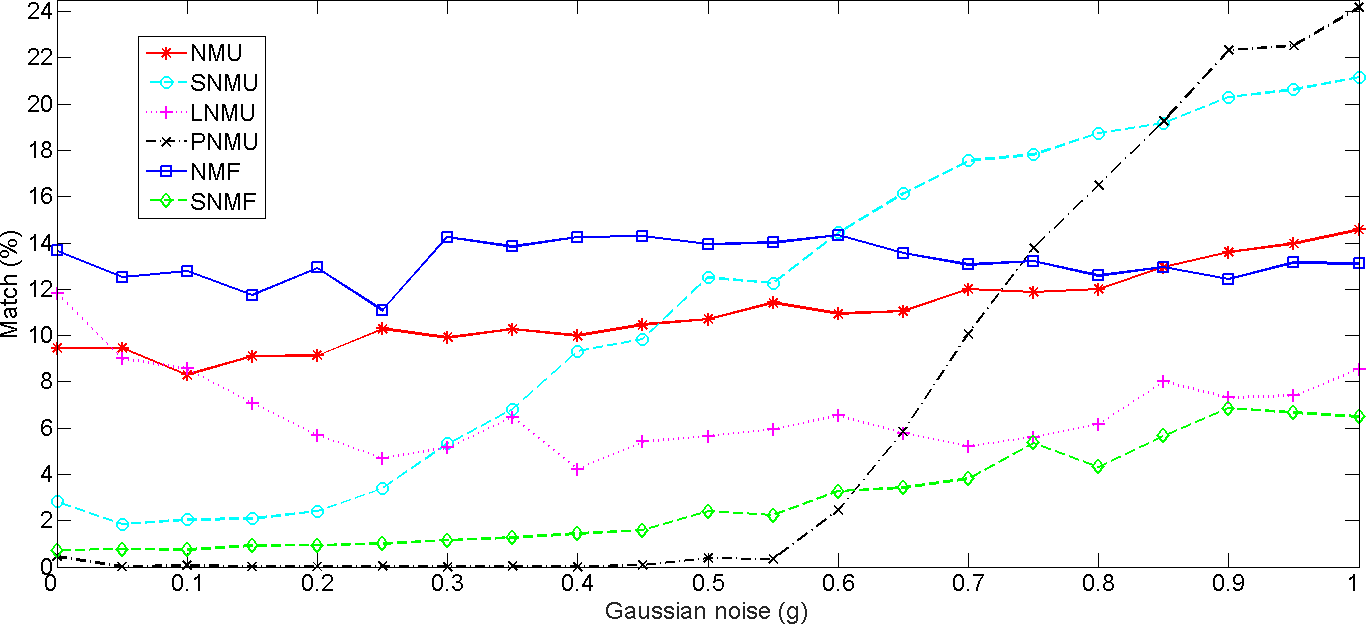} 
				\caption{ Evolution of the average match value~\eqref{match} over 20 different synthetic data sets for different values of the Gaussian noise level ($g$) and for a fixed salt-and-pepper noise level ($p=5\%$). \label{gnnoise}}
 	\end{subfigure}%
	\\ \vspace{0.2cm} 
 	\begin{subfigure}[b]{0.9\textwidth}
 		\includegraphics[width=0.9\textwidth]{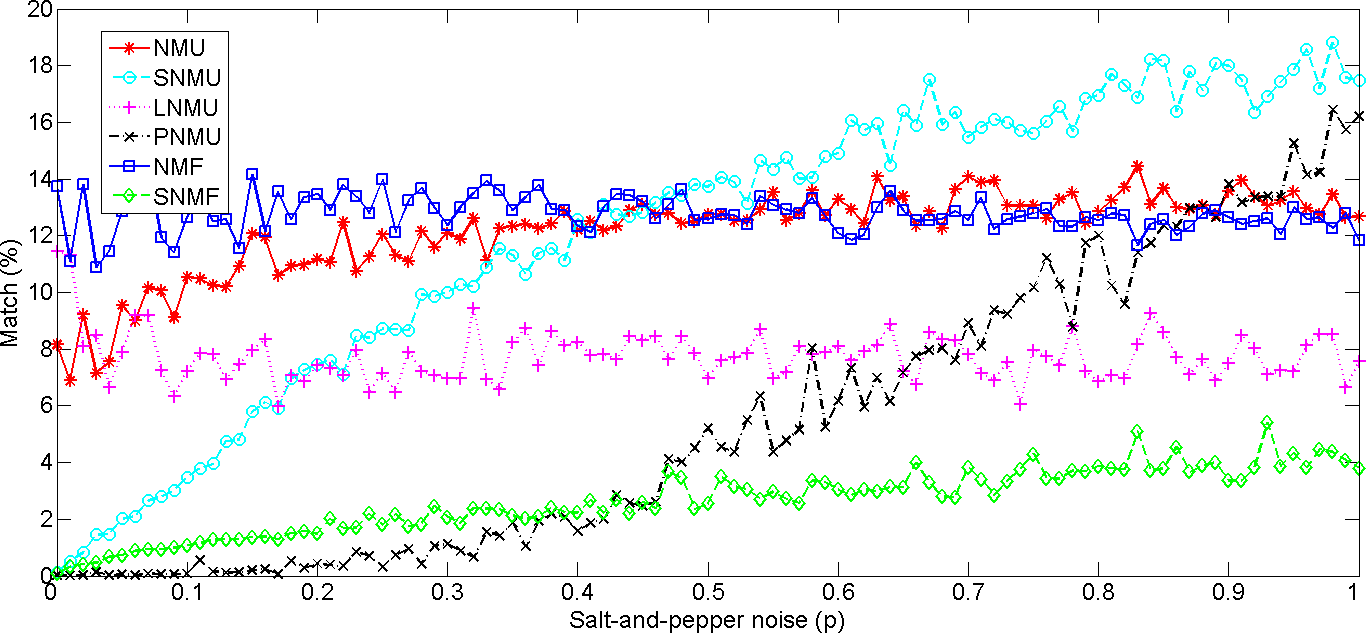} 
				\caption{ Evolution of the average match value~\eqref{match} over 20 different synthetic data sets for different values of the salt-and-pepper noise level ($p$) and for a fixed  the Gaussian noise level ($g=10\%$). \label{psnoise}}
 	\end{subfigure} 
	\\ \vspace{0.2cm} 
 	\begin{subfigure}[b]{0.9\textwidth}
 		\includegraphics[width=0.9\textwidth]{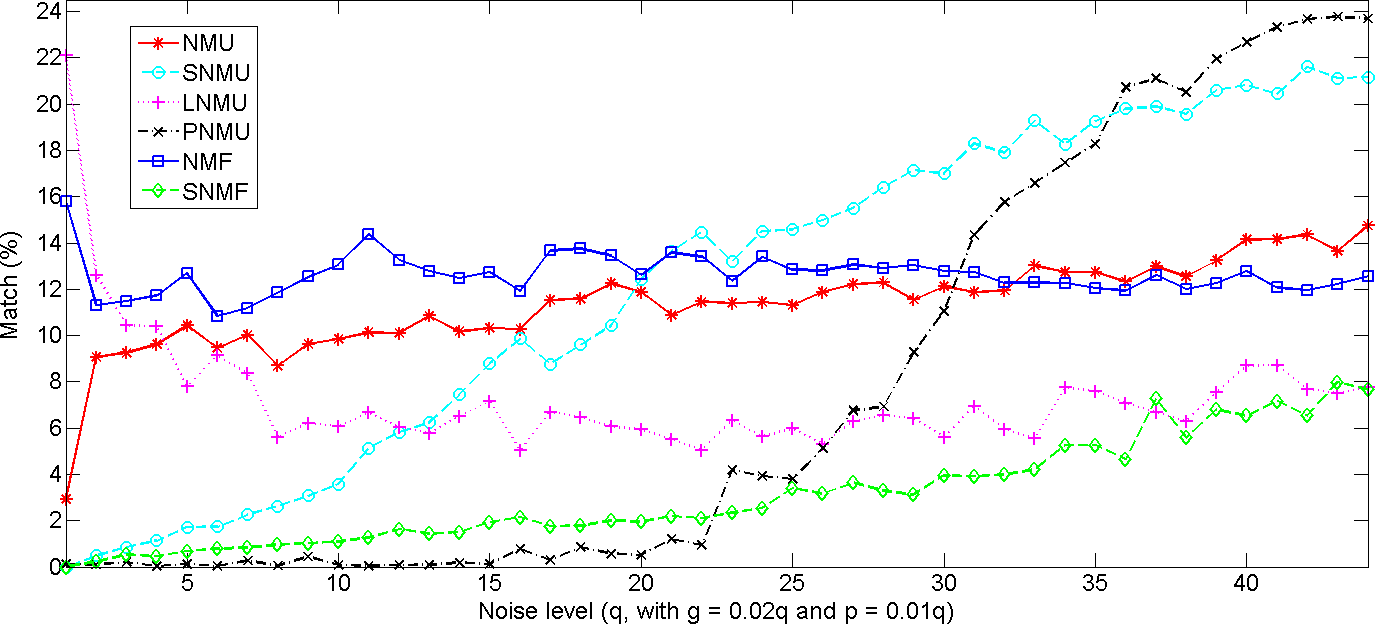} 
				\caption{ Evolution of the average match value~\eqref{match} over 20 different synthetic data sets for different values of the Gaussian noise level ($g=0.02q$) and the salt-and-pepper noise level 
				($p=0.01q$). \label{gnpsnoise}}
 	\end{subfigure}%
 \end{figure}

Let us vary the noise level in three different ways: 
\begin{itemize} 

\item We fix $p=5\%$ and we vary $g = 0,0.05,0.1,\dots,1$. 
Figure~\ref{gnnoise} shows the average value of the match in percent for the different algorithms. We observe that for all values of $g$ smaller than $55\%$, PNMU outperforms the other approaches, being able to generate basis elements with match much smaller than 0.5\% 
(in average, 0.12\%). 
For higher noise levels, the performance of PNMU degrades rapidly.  
SNMF performs relatively well although the match is always higher than 0.7\%. For high noise levels, it performs better than PNMU although for these values of the noise, all basis elements generated are relatively poor.  

 The other approaches perform rather poorly, with most of the match values higher than 5\%. 

\item We fix $g=10\%$ and we use $p = 0,0.01,\dots,1$. 
Figure~\ref{psnoise} shows the average value of the match in percent for the different algorithms. PNMU performs best up to $p=35\%$ (which is rather high as 35\% of the entries of $M$ are highly perturbed) with match smaller than $0.15\%$ for all $p \leq 0.1$, 
and average match of $0.22\%$ for $0.1 \leq p \leq 0.2$. 
It is followed by SNMF while the other approaches perform relatively poorly. 

\item We use $g = 0.02 \, q$ and $p = 0.01 \, q$ with $q = 0,1,2,\dots,50$.  
Figure~\ref{gnpsnoise} shows the average value of the match in percent for the different algorithms. Up to $g=20$, PNMU performs best with average match values smaller than $1\%$. 
For $g \geq 23$ which is rather high ($g = 46\%$ and $p = 23\%$), PNMU deteriorates rapidly but, although SNMF has the lowest match values, they are relatively high and correspond to poor basis elements (and we see that LNMU has similar match values for these high noise levels). 

\end{itemize} 

In all cases, we observe that PNMU outperforms the other NMF and NMU variants where the noise regime is reasonable. 
In particular, it performs very well (match $< 1\%$) for values up to 
$g=0.1,p=0.3$ (Figure~\ref{psnoise}),  
$g=0.55,p=0.05$ (Figure~\ref{gnpsnoise}) and 
$g=0.4,p=0.2$ (Figure~\ref{gnnoise}), 
which is not the case for the other algorithms.

\subsection{Real-world data sets} 

In this section, we first illustrate the sensitivity of PNMU with respect to the parameters $\mu'$ and $\varphi'$ on the Cuprite hyperspectral image.   
Not surprisingly, as we have already observed in the previous section, these parameters play a critical role. 
However, as opposed to classical NMU that is completely unsupervised, NMU with prior information may require human supervision for tuning its parameters and choosing a good trade-off between the reconstruction accuracy, and the spatial coherence and sparsity of the features. 

Then, a quantitative analysis is conducted. Because the ground truth is not know for these data sets, we use three measures: the relative error in percent 
\[ 
\text{Relative Error} = 100\frac{\left\Vert M-UV^{\mathrm{T}}\right\Vert _{F}}{\left\Vert M\right\Vert _{F}}, 
\] 
the sparsity (percentage of nonzero entries) of the factors $U$ 
\[
s\left(U\right)=100\frac{\#\text{zeros}\left(U\right)}{m r} \in \left[0,100\right] 
\; \text{ for } U \in \mathbb{R} ^{m \times r}, 
\] 
and the spatial coherence of the basis elements 
\[
\ell(U) = \sum_{k = 1}^r \frac{|| N U(:,k) ||_1}{ ||U(:,k)||_2} . 
\]
Recall that $|| N U(:,k) ||_1$ amounts for the spatial coherence of the $k$th column of the abundance matrix $U$, while we normalize the columns of $U$ to have a fair comparison (since the algorithms do not normalize the columns of $U$ in the same way). 
The aim of our experiments is to compare the algorithms in terms of the trade-off between these three measures.


Recall that $|| N U(:,k) ||_1$ amounts for the spatial coherence of the $k$th column of the abundance matrix $U$, while we normalize the columns of $U$ to have a fair comparison (since the algorithms do not normalize the columns of $U$ in the same way). 
The aim of our experiments is to compare the algorithms in terms of the trade-off between these three measures. 

Adding constraint to the factorization process leads to an increase of the approximation error, 
but the constrained variants return sparser and/or spatially more coherent solutions. 
As already noted in~\cite{GG09}, it is not fair to compare directly the approximation error of NMU with NMF because NMU computes a solution sequentially. In order to compare the quality of the generated sparsity patterns, one can postprocess the solutions obtained by all algorithms by optimizing over the non-zero entries of $U$ and $V$ (A-HALS can be easily adapted to handle this situation). 
We will refer to \emph{``Improved''} as the relative approximation after this post-processing step. 

Two sets of experimentation are conducted.  The aim of the first experiments is to show the behavior of the proposed method, and to test its effectiveness in correctly detecting the endmembers in a widely used hyperspectral image, the Cuprite image, and reducing noise in the abundance maps. 
The second experiment shows the effectiveness of PNMU in correctly detecting the parts of facial images on one of the most widely used data set in the NMF literature, namely the MIT-CBCL face data set.

\subsubsection{Cuprite} \label{Cuprite}

The Cuprite data set\footnote{Available at \url{http://speclab.cr.usgs.gov/PAPERS.imspec.evol/aviris.evolution.html}.} is widely used to assess the performance of blind hyperspectral unmixing techniques. 
It represents spectral data collected over a mining area in southern Nevada, Cuprite. 
It consists of 188 images with $250\times191$ pixels containing about 20 different endmembers (minerals), although it is unclear how many minerals are present and where they are located precisely; see, e.g.,~\cite{ND05, ACM11} for more information. 
First we discuss the influence of the parameters $\mu'$ and $\varphi'$ on the abundances obtained with PNMU. Then a quantitative comparison is performed to show the effectiveness of PNMU compared to the tested NMU and NMF variants. 


\paragraph{Sensitivity of PNMU to the Parameters $\mu'$ and $\varphi'$}  \label{cupsens}

\begin{figure}
 	\centering
 	\begin{subfigure}[b]{\textwidth}
 		\includegraphics[width=1\textwidth]{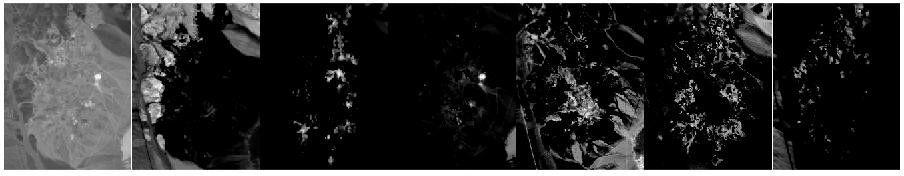} 
				\caption{PNMU with parameters $\mu'=0.1$ and $\varphi'=0.2$ \label{Cuprite_mu01_phi02}.}
 	\end{subfigure}%
	\\ \vspace{0.2cm} 
 	\begin{subfigure}[b]{\textwidth}
 		\includegraphics[width=1\textwidth]{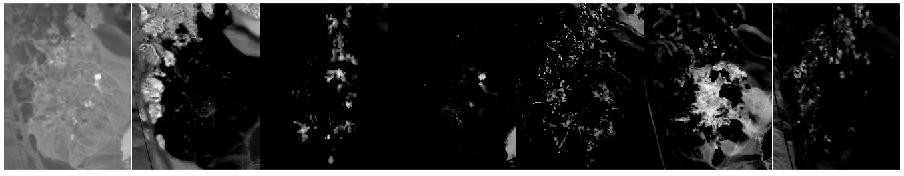} 
				\caption{PNMU with parameters $\mu'=0.3$ and $\varphi'=0.2$\label{Cuprite_mu03_phi02}.}
 	\end{subfigure} 
	\\ \vspace{0.2cm} 
 	\begin{subfigure}[b]{\textwidth}
 		\includegraphics[width=1\textwidth]{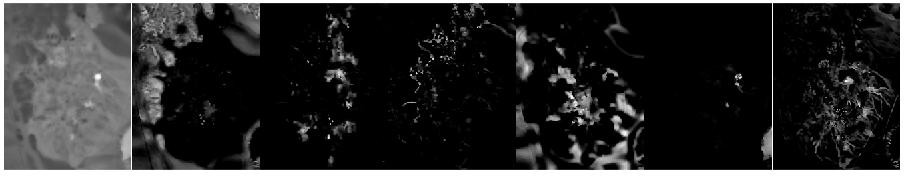}
		\caption{PNMU with parameters $\mu'=0.5$ and $\varphi'=0.2$\label{Cuprite_mu05_phi02}.}
 	\end{subfigure}%
 	\\ \vspace{0.2cm} 
 	\begin{subfigure}[b]{\textwidth}
 		\centering
 		\includegraphics[width=1\textwidth]{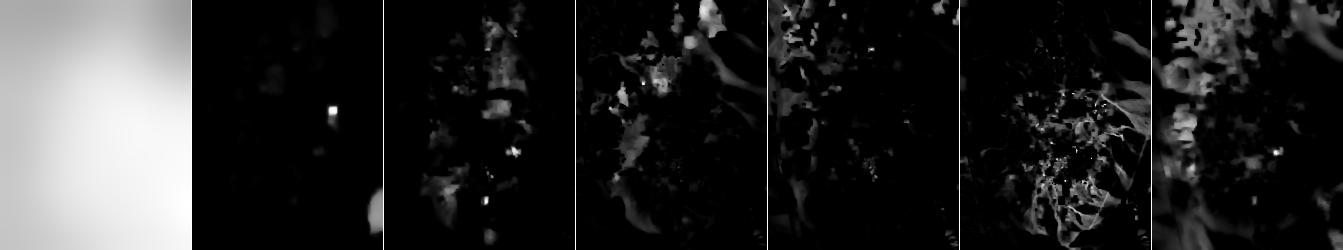}
 		\caption{PNMU with parameters $\mu'=0.7$ and $\varphi'=0.2$\label{Cuprite_mu07_phi02}.}
 	\end{subfigure}%
 	\\ \vspace{0.2cm} 
 	\begin{subfigure}[b]{\textwidth}
 		\centering
 		\includegraphics[width=1\textwidth]{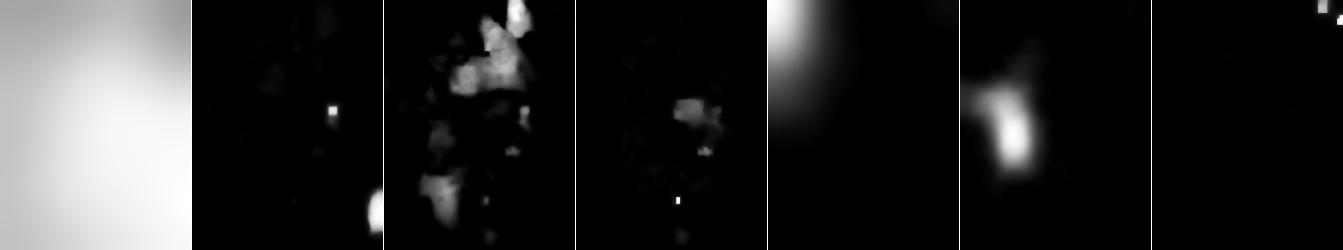}
 		\caption{ PNMU with parameters $\mu'=0.9$ and $\varphi'=0.2$\label{Cuprite_mu09_phi02}.}
 	\end{subfigure}%
 	\caption{Influence of the locality term on the first seven abundance images for the Cuprite data set. \label{comparison_locality}}
 \end{figure}

In this section, we conduct some experiments to show the sensitivity of PNMU to the parameters $\mu'$ and $\varphi'$. 
To show a wide range of abundance elements for different values of $\mu'$ and $\varphi'$, we only extract 7 abundance columns ($r = 7$, light tones in the figures indicate a high degree of membership). 
 Figure \ref{comparison_locality} shows the first seven abundance images extracted by PNMU, 
when the penalty for sparsity is fixed ($\varphi'= 0.2$) and the penalty for spatial coherence varies ($\mu'= 0.1,0.3,\dots,0.9$).
 We observe that increasing $\mu'$ improves the spatial coherence of the abundance images. 
However, high values of $\mu'$ (starting for $\mu' = 0.5$) lead to blurred images and lost of contours (sub-figures~\ref{Cuprite_mu05_phi02}, \ref{Cuprite_mu07_phi02} and \ref{Cuprite_mu09_phi02}). 
 
\begin{figure}
 	\centering
 	\begin{subfigure}[b]{\textwidth}
 		\centering
 		\includegraphics[width=1\textwidth]{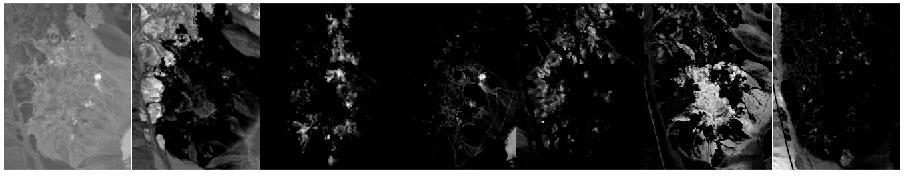}
 		\caption{PNMU with parameters $\mu'=0.1$ and $\varphi'=0.1$.}
 	\end{subfigure}%
 	\\ \vspace{0.2cm} 
 	\begin{subfigure}[b]{\textwidth}
 		\centering
 		\includegraphics[width=1\textwidth]{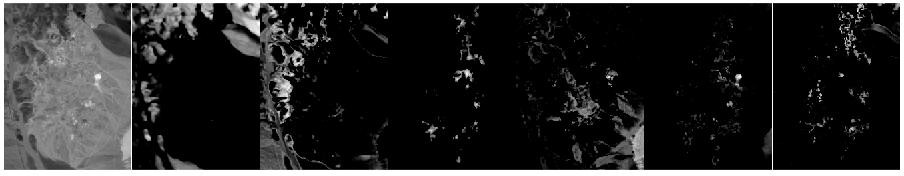}
 		\caption{PNMU with parameters $\mu'=0.1$ and $\varphi'=0.3$.}
 	\end{subfigure}%
 	\\ \vspace{0.2cm} 
 	\begin{subfigure}[b]{\textwidth}
 		\centering
 		\includegraphics[width=1\textwidth]{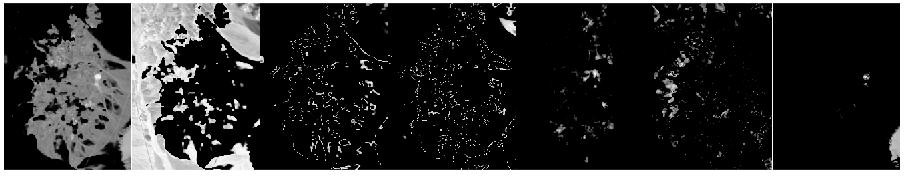}
 		\caption{PNMU with parameters $\mu'=0.1$ and $\varphi'=0.5$. \label{Cuprite_mu01_phi05}}
 	\end{subfigure}%
 	\\ \vspace{0.2cm} 
 	\begin{subfigure}[b]{\textwidth}
 		\centering
 		\includegraphics[width=1\textwidth]{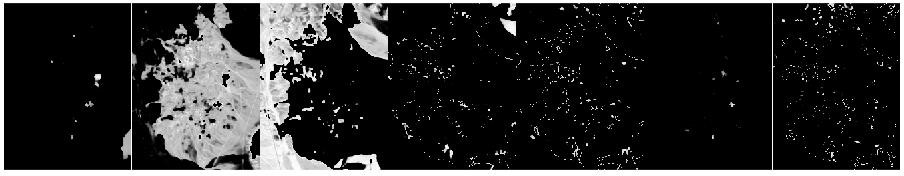}
 		\caption{PNMU with parameters $\mu'=0.1$ and $\varphi'=0.7$. \label{Cuprite_mu01_phi07}}
 	\end{subfigure}%
 	\\ \vspace{0.2cm} 
 	\begin{subfigure}[b]{\textwidth}
 		\centering
 		\includegraphics[width=1\textwidth]{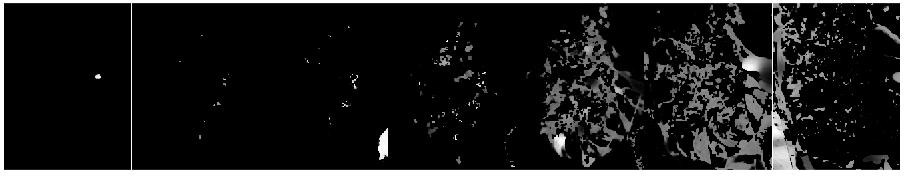}
 		\caption{PNMU with parameters $\mu'=0.1$ and $\varphi'=0.9$ \label{Cuprite_mu01_phi09}.}
 	\end{subfigure}%
 	\caption{Influence of the sparsity term on the first seven abundance images for the Cuprite data set. \label{comparison_sparsity}}
 \end{figure}

Adding sparsity constraints into NMU allows to detect endmembers more effectively because sparsity prevents the mixture of different endmembers in one abundance element~\cite{GP11}. 
 Figure~\ref{comparison_sparsity} shows the first seven bases obtained with PNMU for $\mu'$ fixed to $0.1$ and 
where $\varphi'$ varies ($\varphi'= 0.1,0.3,\dots,0.9$). 
It can be observed that, as for the spatial information, 
small values of the sparsity parameter $\varphi'$ ($0.1$ or $0.3$) provide good results; in fact, for higher values of this parameter, the abundances are too sparse, and PNMU is not able to detect good features; see sub-figures~\ref{Cuprite_mu05_phi02}, \ref{Cuprite_mu07_phi02} and \ref{Cuprite_mu09_phi02}. 
 
 Furthermore, it is necessary to point out that the sparsity and the locality constraints influence each other. 
Indeed, higher sparsity implies that less pixels are present in each abundance element which improves the spatial coherence (at the limit, all pixels take the value zero and the spatial coherence is perfect). For example, we observe on Figure~\ref{comparison_locality} that increasing $\mu'$ also strongly influences sparsity. 
A human supervision is necessary to adequately tune these parameters. 
For the Cuprite data set, we have observed that PNMU with  $\mu'=0.1$ and $\varphi'=0.2$ returns basis elements with a good tradeoff between sparsity and spatial coherence.

\paragraph{Quantitative Comparison} 

\begin{figure}
	\centering
	\begin{subfigure}[b]{\textwidth}
		\centering
		\includegraphics[width=.7\textwidth]{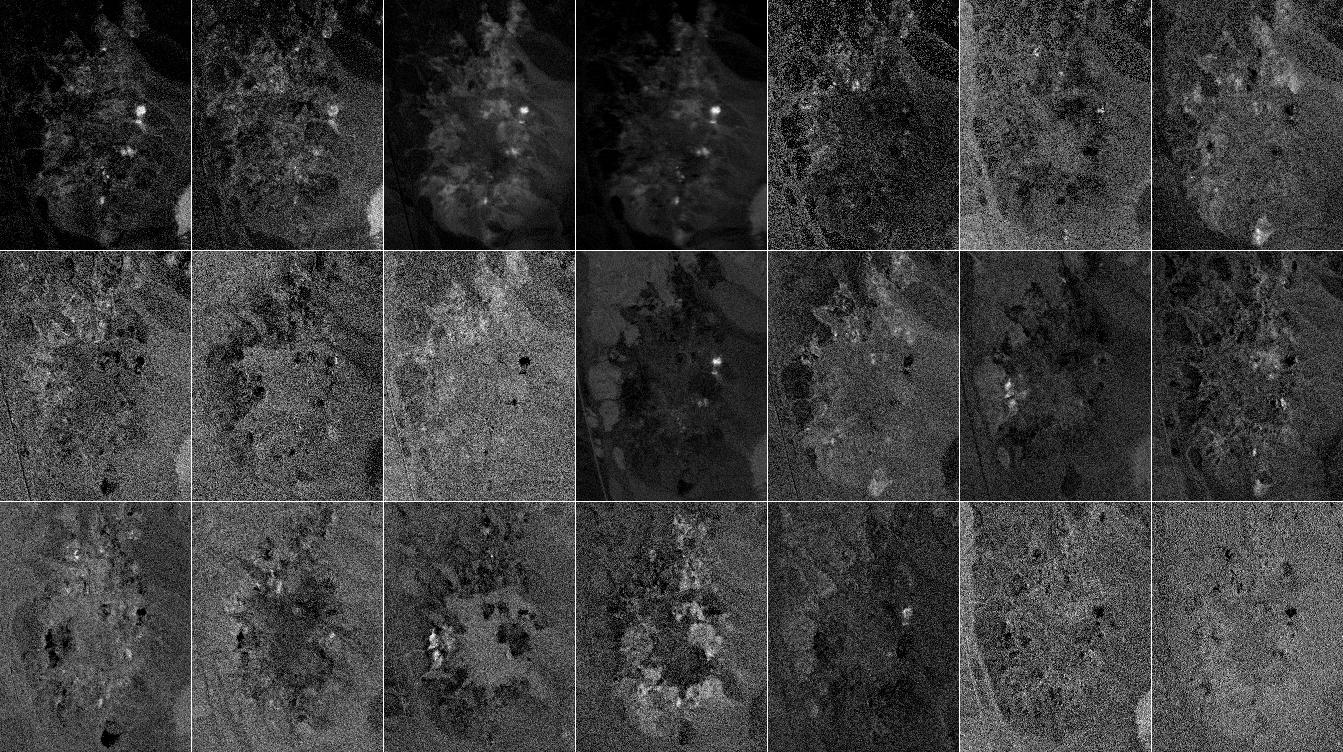}
		\caption{NMF.\label{Cuprite_NMF}}
	\end{subfigure}
	\\
	\begin{subfigure}[b]{\textwidth}
		\centering
		\includegraphics[width=.7\textwidth]{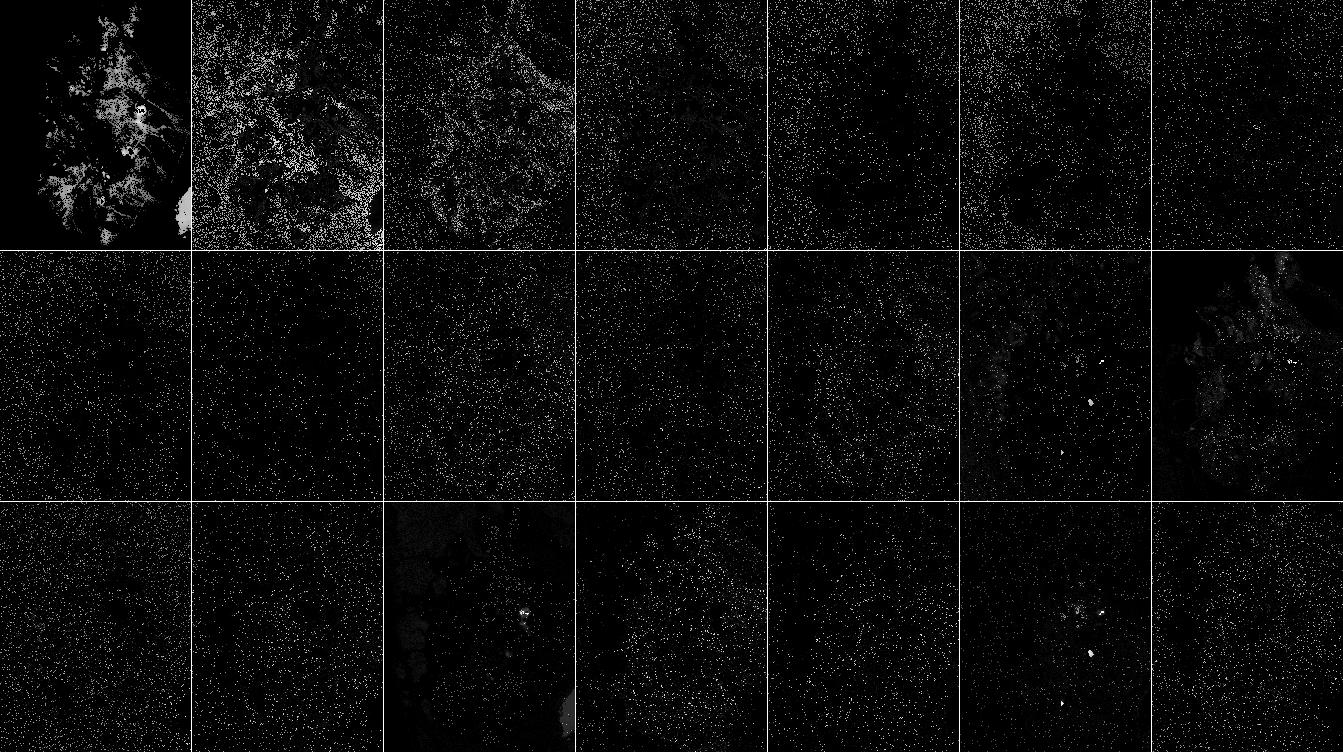}
		\caption{SNMF. \label{Cuprite_SNMF}}
	\end{subfigure}
	\\ 
	\begin{subfigure}[b]{\textwidth}
		\centering
		\includegraphics[width=.7\textwidth]{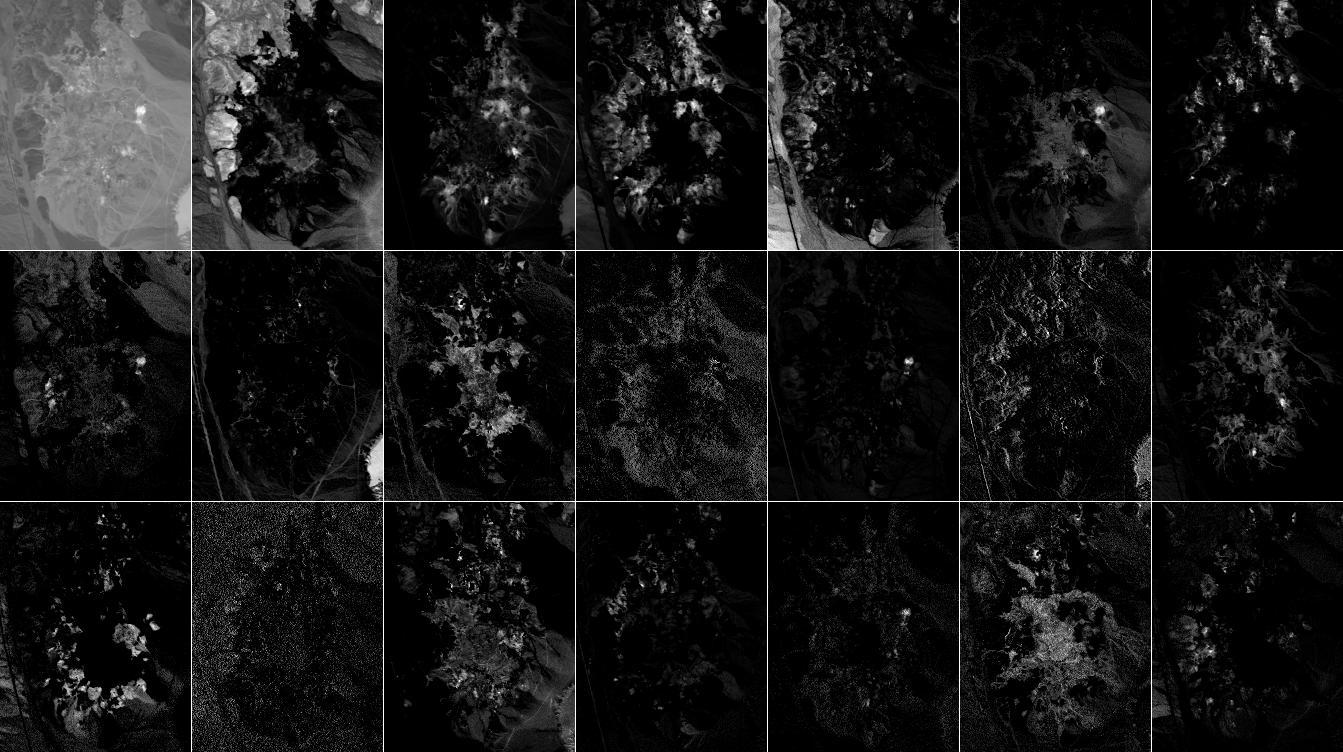}
		\caption{NMU
		.\label{Cuprite_NMU}}
	\end{subfigure}%
	\caption{Abundance elements obtained for the Cuprite data set with NMF, SNMF and NMU algorithms.\label{Cuprite_comparison1}}
\end{figure}

\begin{figure}
	\centering
	\begin{subfigure}[b]{\textwidth}
		\centering
		\includegraphics[width=.7\textwidth]{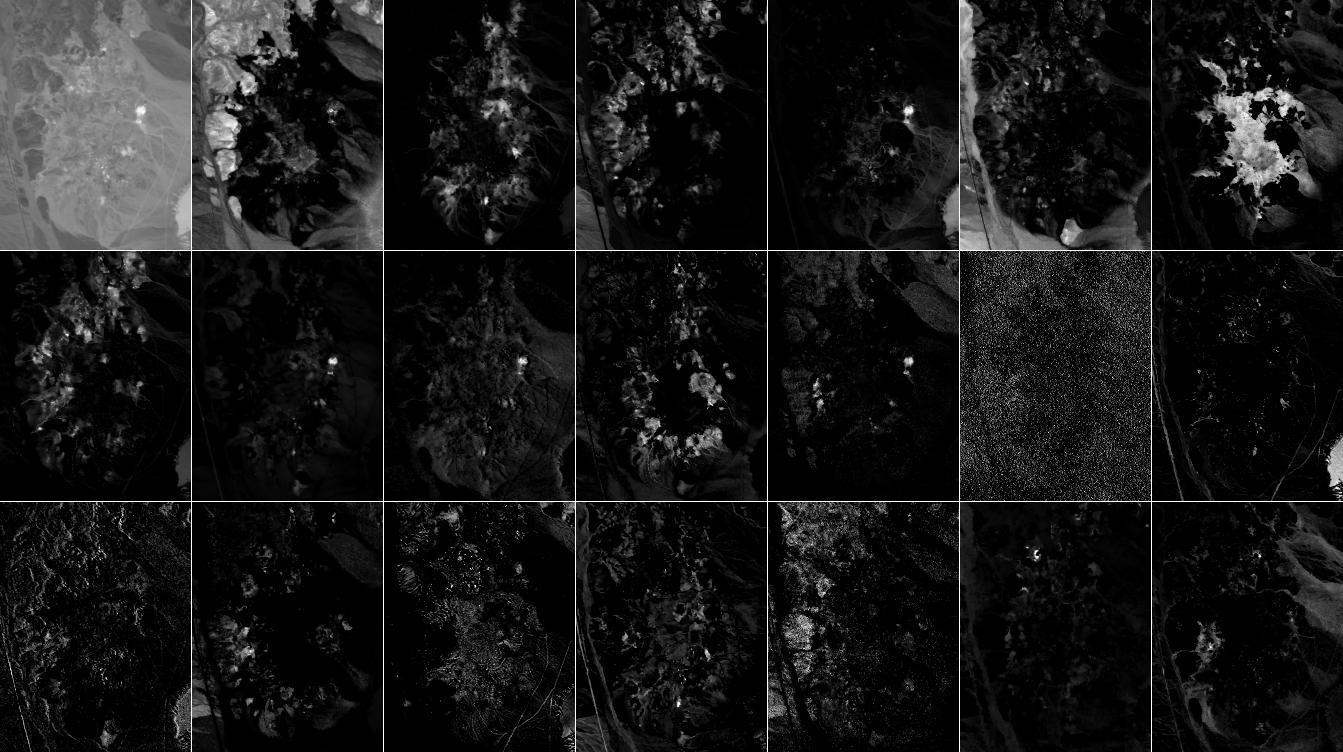}
		\caption{LNMU
			. \label{Cuprite_LNMU}}
	\end{subfigure}
	\\ 
	\begin{subfigure}[b]{\textwidth}
		\centering
		\includegraphics[width=.7\textwidth]{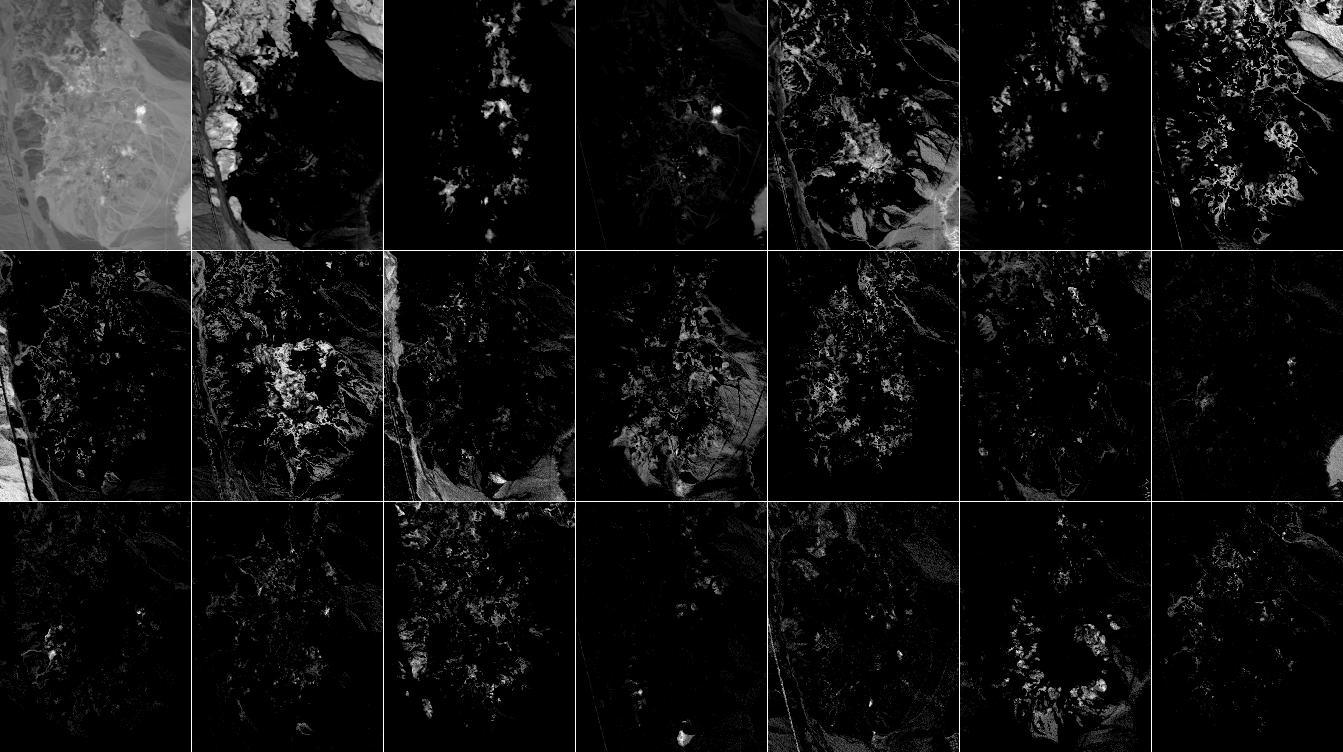}
		\caption{SNMU.\label{Cuprite_SNMU}}
	\end{subfigure}
	\\
	\begin{subfigure}[b]{\textwidth}
		\centering
		\includegraphics[width=.7\textwidth]{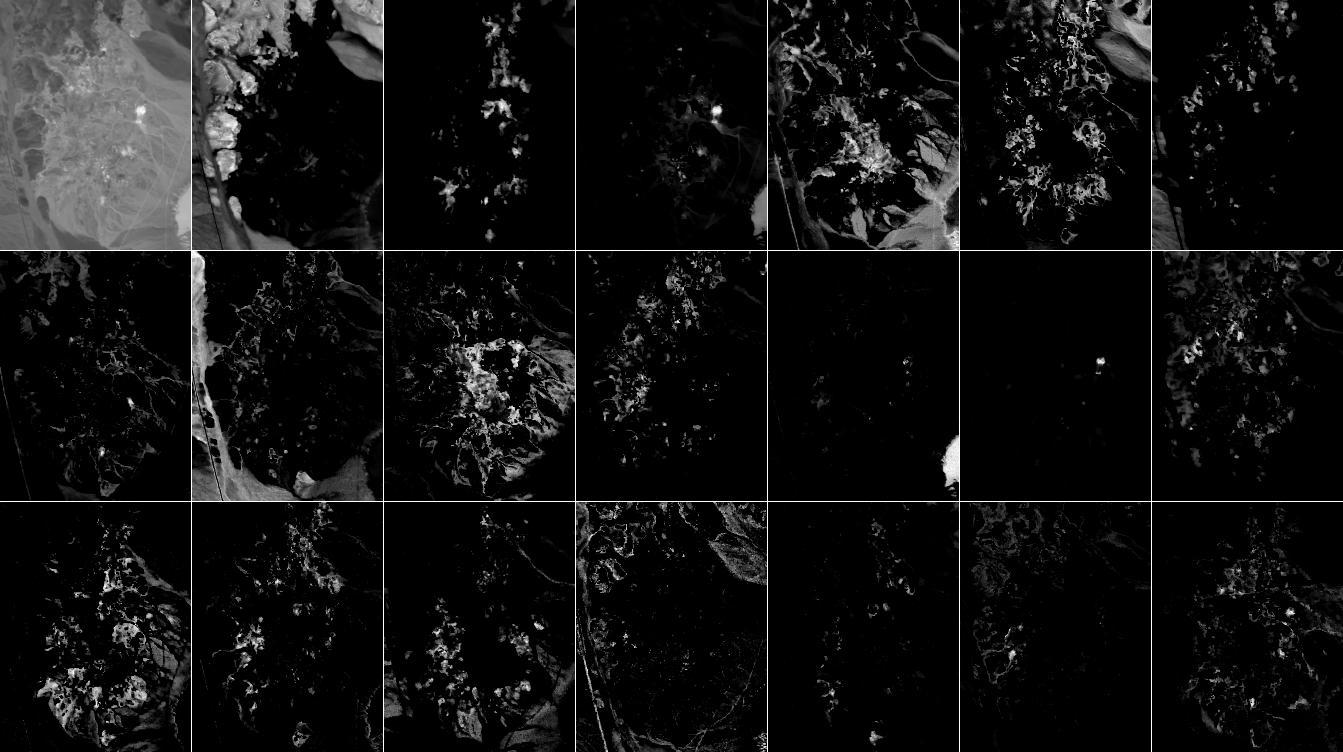}
		\caption{PNMU. \label{Cuprite_PNMU}}
	\end{subfigure}%
	\caption {Abundance elements obtained for the Cuprite data set with LNMU, SNMU and PNMU.\label{Cuprite_comparison2}}
\end{figure}

We now compare PNMU with the algorithms listed in the introduction of Section~\ref{expres} to show its ability to extract features with a good trade-off between reconstruction error, sparsity and spatial coherence.   
Figures~\ref{Cuprite_comparison1} and~\ref{Cuprite_comparison2} display the abundance images obtained by the different algorithms for $r = 21$ on the Cuprite data set.  
Table~\ref{Cuprite_table} provides a quantitative comparison of the different algorithms, with the values of the relative error, sparsity and spatial coherence. 
\begin{table}[ht]
	\centering
	\begin{tabular}{|c|c|c|c|c|c|c|}
		\hline  & NMF & SNMF & NMU & LNMU & SNMU & PNMU   \\ 
		\hline Error &  0.62  &  2.59 & 1.37 & 1.44 & 1.87  &  1.85   \\ 
		\hline Improved &  \textbf{0.61} &  2.02 &   0.71 & 0.66 & 1.13 & 1.09   \\ 
		\hline $s\left(U\right)$ & 3.76 &  \textbf{77.96} &  50.41 & 40.90 & 76.33 & 75.29     \\ 
\hline $\ell\left( U \right)$ & 3606 &   3829  & 2585 & 2039 & 2292 &  \textbf{1381}   \\ 
		\hline 
	\end{tabular}
	\caption{Comparison of the relative approximation error, the sparsity and the spatial coherence for the Cuprite data set.\label{Cuprite_table}}
\end{table}

We observe the following:
\begin{itemize}

\item NMF is not able to detect endmembers, and generates mostly dense hence not localized abundance images, 
most of them containing salt-and-pepper-like noise; see sub-figure  \ref{Cuprite_NMF}. 

This qualitative observation is confirmed by the quantitative measurements from Table~\ref{Cuprite_table}: 
although NMF has the lowest reconstruction error (since it only focuses on minimizing it), 
it has the lowest value for the sparsity of the abundances, and the second highest value for their spatial coherence 
(higher values mean worse spatial coherence).

\item Despite the fact that SNMF gives sparser solutions than NMF, the abundance images are surprisingly very poor in terms of local coherence (highest value); see sub-figure \ref{Cuprite_SNMF}. 

\item NMU returns sparse abundances, and is able to localize different endmembers (see the description of the PNMU abundance maps below), even if some images are rather noisy and the edges defining the materials are not always well identified; see, e.g., 11th and 16th abundance elements in sub-figure \ref{Cuprite_NMU}. 
It generates sparser and spatially more coherent features than NMF, while its relative error is comparable (0.61\% vs.\@ 0.71\%). 

\item Not surprisingly, LNMU provides spatially more coherent (from 2585 to 2039) 
but denser (from 50.41 to 40.90) bases than NMU. In particular, the edge delimitation of several materials is better defined in the LNMU abundance maps, although some are still rather noisy (e.g., 13th, 17th, 19th) 
and relatively dense (e.g., 6th, 8th); see sub-figure \ref{Cuprite_LNMU}. 

\item Adding the penalty term inducing sparsity on NMU allows to remove some of the noise, 
though the edges of some materials are still not well defined (sub-figure \ref{Cuprite_SNMU}). 
In fact, compared to NMU, SNMU quite naturally increases the sparsity of NMU abundances (from $50.41$ to $76.33$) and improves the spatial coherence (from 2585 to 2292), although the relative error is slightly increased (from 0.71\% to 1.13\%). 
  

\item PNMU achieves the best trade-off between relative error 
(1.1\% vs.\@ 0.6\% for NMF hence an increase of only 0.5\%, while NMF clearly generates poor basis elements because they are mostly dense and noisy--see the discussion above), sparsity (close to that of SNMF, 77.96\% vs.\@ 75.29\%), 
and spatial coherence (PNMU achieves the lowest value). 
Sub-figure~\ref{Cuprite_PNMU} shows the abundance images obtained with PNMU for the Cuprite data set with parameters $\mu'=0.1$ and $\varphi'=0.2$. 
It can be observed that the images are less noisy (as for SNMU), and the edges of materials are well defined (as for LNMU). 
Hence, PNMU combines the advantages of all methods on this data set. 
In fact, it is able to identify many endmembers (see~\cite{ND05, ACM11} where these materials are identified) among which, from left to right, top to bottom (sub-figure~\ref{Cuprite_PNMU}): 
		(2)~Desert Varnish, 
		(3)~Alunite~1, 
		(5)~Chalcedony and Hematite, 
		(6)~Alunite~2, 
		(7)~Kaolinite~1 and Goethite, 
		(9)~Amorphous iron oxydes and Goethite,
		(10)~Chalcedony, 
		(12)~Montmorillonite, 
		(13)~Muscovite, 
		(14)~Kaolinite~1, 
		(15)~K-Alunite, 
		(18)~Kaolonite~2, and 
		(20)~Buddingtonite. 
\end{itemize}

\subsubsection{MIT-CBCL Face database}
\label{CBCL}

The previous experiments show the effectiveness of the proposed method in identifying materials that compose a hyperspectral image. 
We now apply PNMU on MIT-CBCL Face database\footnote{\url{http://cbcl.mit.edu/software-data sets/FaceData2.html} 
} with 2429 faces, $19\times19$ pixels each. This is one of the most widely used data set in the NMF literature and was used in the original paper of Lee and Seung~\cite{LS1}. 
This is a database of faces and non-faces images, used at the Center for Biological and Computational Learning at MIT and we use the faces subset of the training set.

We use this data set to show that PNMU can also be used for other types of images, and provides meaningful and useful results compared to NMF and NMU variants.  As for the Cuprite data set, we will use $\mu'=0.1$ and $\varphi'=0.2$. 

\begin{figure}
	\centering
	\begin{subfigure}[b]{0.45\textwidth}
		\centering
		\includegraphics[width=0.80\textwidth]{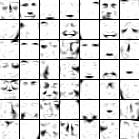}
		\caption{NMF.\label{CBCL_NMF}}
	\end{subfigure}%
	~ 
	\begin{subfigure}[b]{0.45\textwidth}
		\centering
		\includegraphics[width=.80\textwidth]{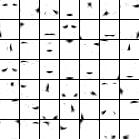}
		\caption{SNMF. \label{CBCL_SNMF}}
	\end{subfigure}%
	\\ \vspace{0.2cm} 
	\begin{subfigure}[b]{0.45\textwidth}
		\centering
		\includegraphics[width=.80\textwidth]{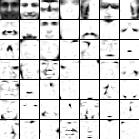}
		\caption{NMU
			.\label{CBCL_NMU}}
	\end{subfigure}%
	~ 
	\begin{subfigure}[b]{0.45\textwidth}
		\centering
		\includegraphics[width=.80\textwidth]{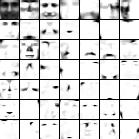}
		\caption{LNMU
			.\label{CBCL_LNMU}}
	\end{subfigure}%
	\\ \vspace{0.2cm} 
	\begin{subfigure}[b]{0.45\textwidth}
		\centering
		\includegraphics[width=.80\textwidth]{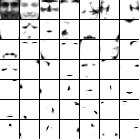}
		\caption{SNMU
			.\label{CBCL_SNMU}}
	\end{subfigure}%
	 ~
	\begin{subfigure}[b]{0.45\textwidth}
		\centering
		\includegraphics[width=.80\textwidth]{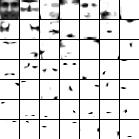}
		\caption{PNMU
			. \label{CBCL_PNMU}}
	\end{subfigure}%
	\caption{Abundance elements obtained for the MIT-CBCL Face data set with the different algorithms.\label{CBCL_comparison}}
\end{figure}

Figure~\ref{CBCL_comparison} displays the abundance images obtained with the different algorithms, while Table~\ref{CBCL_table} reports the numerical results. As for hyperspectal images, PNMU provides a good trade-off between reconstruction error, sparsity and spatial coherence. 
\begin{table}[ht] 
	\centering
	\begin{tabular}{|c|c|c|c|c|c|c|}
		\hline  & NMF & SNMF  & NMU & LNMU & SNMU & PNMU   \\ 
		\hline Error &  8.18 &  8.67 &  15.29 & 15.90 & 15.50 & 14.63    \\ 
		\hline Improved & \textbf{8.16} &    8.46 &  8.76 & 8.46 & 9.80 &  9.52   \\ 
		\hline $s\left(U\right)$ & 54.89 &  84.80 & 70.05 & 58.65 & \textbf{87.69} &  85.77    \\ 
		\hline $\ell\left(U\right)$ & 457 &  310 &  419  & \textbf{55} & 334 & 271      \\
		\hline 
	\end{tabular}
	\caption{Comparison of the relative approximation error, the sparsity and the spatial coherence for the MIT-CBCL data set.\label{CBCL_table}}
\end{table} 

NMF, NMU and LNMU generate mixed abundances where different parts of the faces are represented (for example in the sub-figure \ref{CBCL_NMU}, the third abundance image in the fourth row, includes the nose, the mouth, the eyes and the eyebrows all together). The sparse methods (SNMU, PNMU, SNMF) are able to detect unmixed parts of the faces, but among these, the parts returned by PNMU are more localized. 
In fact, PNMU has a sparsity level comparable with the other methods enhancing sparsity, but has a better spatial coherence. 


\section{Conclusions} 

In this paper a variant of NMU was proposed, namely PNMU, taking into account both sparsity and spatial coherence of the abundance elements.   
Numerical experiments have shown the effectiveness of PNMU in correctly generating sparse and localized features in images (in particular, synthetic, hyperspectral and facial images), with a better trade-off between sparsity, spatial coherence and reconstruction error. 

\bibliographystyle{siam}
\bibliography{PNMU}

\begin{thebibliography}{10}

\bibitem{ACM11}
{\sc A.~Ambikapathi, T.-H. Chan, W.-K. Ma, and C.-Y. Chi}, {\em
  Chance-constrained robust minimum-volume enclosing simplex algorithm for
  hyperspectral unmixing}, IEEE Transactions on Geoscience and Remote Sensing,
  49 (2011), pp.~4194--4209.

\bibitem{AW07}
{\sc K.~M. Anstreicher and L.~A. Wolsey}, {\em {Two "well-known" properties of
  subgradient optimization}}, Mathematical Programming, 120(1) (2009),
  pp.~213--220.

\bibitem{BGG09}
{\sc M.W. Berry, N.~Gillis, and F.~Glineur}, {\em {Document Classification
  Using Nonnegative Matrix Factorization and Underapproximation}}, in Proc. of
  the IEEE Int.\@ Symp.\@ on Circuits and Systems (ISCAS), 2009,
  pp.~2782--2785.

\bibitem{CAZP09}
{\sc A.~Cichocki, S.~Amari, R.~Zdunek, and A.H. Phan}, {\em Non-negative Matrix
  and Tensor Factorizations: Applications to Exploratory Multi-way Data
  Analysis and Blind Source Separation}, Wiley-Blackwell, 2009.

\bibitem{DDF10}
{\sc I.~Daubechies, R.~DeVore, M.~Fornasier, and C.S. G{\"u}nt{\"u}rk}, {\em
  Iteratively reweighted least squares minimization for sparse recovery},
  Communications on Pure and Applied Mathematics, 63 (2010), pp.~1--38.

\bibitem{G12}
{\sc N.~Gillis}, {\em {Sparse and unique nonnegative matrix factorization
  through data preprocessing}}, Journal of Machine Learning Research, 13
  (2012), pp.~3349--3386.

\bibitem{G14}
\leavevmode\vrule height 2pt depth -1.6pt width 23pt, {\em The why and how of
  nonnegative matrix factorization}, in Regularization, Optimization, Kernels,
  and Support Vector Machines, J.A.K. Suykens, M.~Signoretto, and A.~Argyriou,
  eds., Chapman \& Hall/CRC, Machine Learning and Pattern Recognition Series,
  2014, pp.~257--291.

\bibitem{GG09}
{\sc N.~Gillis and F.~Glineur}, {\em Using underapproximations for sparse
  nonnegative matrix factorization}, Pattern Recognition, 43 (2010),
  pp.~1676--1687.

\bibitem{GG11}
\leavevmode\vrule height 2pt depth -1.6pt width 23pt, {\em {Accelerated
  Multiplicative Updates and Hierarchical ALS Algorithms for Nonnegative Matrix
  Factorization}}, Neural Computation, 24 (2012), pp.~1085--1105.

\bibitem{GP10}
{\sc N.~Gillis and R.J. Plemmons}, {\em Dimensionality reduction,
  classification, and spectral mixture analysis using nonnegative
  underapproximation}, Optical Engineering, 50 (2011), p.~027001.

\bibitem{GP11}
\leavevmode\vrule height 2pt depth -1.6pt width 23pt, {\em Sparse nonnegative
  matrix underapproximation and its application to hyperspectral image
  analysis}, Linear Algebra and its Applications, 438 (2013), pp.~3991--4007.

\bibitem{PriorsNicolas}
{\sc N.~Gillis, R.J. Plemmons, and Q.~Zhang}, {\em Priors in sparse recursive
  decompositions of hyperspectral images}, in Proc. SPIE, Algorithms Technol.
  Multispectr., Hyperspectr. Ultraspectr. Imagery XVIII, vol.~8390, 2012,
  p.~83901M.

\bibitem{IBP11}
{\sc M.-D. Iordache, J.M. Bioucas-Dias, and A.~Plaza}, {\em Total variation
  regulatization in sparse hyperspectral unmixing}, in Third Worskshop on
  Hyperspectral Image and Signal Processing: Evolution in Remote Sensing
  (WHISPERS), Lisbon, 2011.

\bibitem{J86}
{\sc I.T. Jolliffe}, {\em Principal Component Analysis}, Springer-Verlag, 1986.

\bibitem{Kim}
{\sc H.~Kim and H.~Park}, {\em Sparse non-negative matrix factorizations via
  alternating non-negativity-constrained least squares for microarray data
  analysis}, Bioinformatics, 23 (2007), pp.~1495--1502.

\bibitem{KCJ11}
{\sc I.~Kopriva, X.~Chen, and Y.~Jao}, {\em {Nonlinear Band Expansion and
  Nonnegative Matrix Underapproximation for Unsupervised Segmentation of a
  Liver from a Multi-phase CT image}}, in SPIE Medical Imaging-Image Processing
  Volume 7962, Orlando, 2011.

\bibitem{KC11}
{\sc I.~Kopriva, M.~Hadzija, M.P. Hadzija, M.~Korolija, and A.~Cichocki}, {\em
  Rational variety mapping for contrast-enhanced nonlinear unsupervised
  segmentation of multispectral images of unstained specimen}, The American
  Journal of Pathology, 179 (2011), pp.~547--554.

\bibitem{LS1}
{\sc D.D. Lee and H.S. Seung}, {\em Learning the parts of objects by
  nonnegative matrix factorization}, Nature, 401 (1999), pp.~788--791.

\bibitem{ND05}
{\sc J.M.P. Nascimento and J.M. Bioucas~Dias}, {\em Vertex component analysis:
  a fast algorithm to unmix hyperspectral data}, IEEE Transactions on
  Geoscience and Remote Sensing, 43 (2005), pp.~898--910.

\bibitem{Y04}
{\sc Yu. Nesterov}, {\em Introductory Lectures on Convex Optimization: A Basic
  Course}, Kluwer Academic Publishers, 2004.

\bibitem{hyper_image2}
{\sc R.~Smith}, {\em Introduction to hyperspectral imaging}, Microimages,
  (2006).
\newblock
  \url{http://www.microimages.com/documentation/Tutorials/hyprspec.pdf}.

\bibitem{V09}
{\sc S.A. Vavasis}, {\em On the complexity of nonnegative matrix
  factorization}, SIAM Journal on Optimization, 20 (2009), pp.~1364--1377.

\bibitem{ZKZ07}
{\sc A.~Zymnis, S.-J. Kim, J.~Skaf, M.~Parente, and S.~Boyd}, {\em
  Hyperspectral image unmixing via alternating projected subgradients}, in
  Signals, Systems and Computers, 2007, 2007, pp.~1164 --1168.

\end{thebibliography}

\end{document}